\documentclass[journal]{IEEEtran}

  \usepackage{amsmath}
  \usepackage{amssymb}
  \usepackage[nocompress]{cite}
  \usepackage{graphicx}
  \usepackage{multicol}
  \usepackage{ragged2e}
  \usepackage{booktabs}
  \usepackage{multirow}
  \usepackage[table,xcdraw]{xcolor}
  \usepackage[normalem]{ulem}
  \usepackage{epsfig}
  \usepackage{subcaption}
  \usepackage{caption}
  \usepackage{amsfonts}
  \usepackage{adjustbox}
  \usepackage{bbding}
  \usepackage{makecell}
  \usepackage{arydshln}
  \usepackage{colortbl}
  \usepackage{overpic}
\usepackage[pagebackref=false,
            breaklinks=true,colorlinks,
            citecolor=cyan,linkcolor=red,
            bookmarks=false]{hyperref}

\ifCLASSINFOpdf
\else
\fi
\usepackage[switch]{lineno}

\begin{document}

\title{Efficient Image Super-Resolution with Feature Interaction Weighted Hybrid Network}

\author{Wenjie Li,
        Juncheng Li,
        Guangwei Gao,~\IEEEmembership{Senior Member,~IEEE,}
        Weihong Deng,~\IEEEmembership{Member,~IEEE,}
        Jian Yang,~\IEEEmembership{Member,~IEEE,}
        Guo-Jun Qi,~\IEEEmembership{Fellow,~IEEE}
        and Chia-Wen Lin,~\IEEEmembership{Fellow,~IEEE}
\thanks{This work was supported in part by the foundation of Key Laboratory of Artificial Intelligence of Ministry of Education under Grant AI202404, and in part by the Open Fund Project of Provincial Key Laboratory for Computer Information Processing Technology (Soochow University) under Grant KJS2274.~\textit{(Corresponding author: Guangwei Gao.)}}
\IEEEcompsocitemizethanks{\IEEEcompsocthanksitem Wenjie Li and Weihong Deng are with the Pattern Recognition and Intelligent System Laboratory, School of Artificial Intelligence, Beijing University of Posts and Telecommunications, Beijing 100080, China (e-mail: lewj2408@gmail.com, whdeng@bupt.edu.cn).
\IEEEcompsocthanksitem Guangwei Gao is with the Intelligent Visual Information Perception Laboratory, Institute of Advanced Technology, Nanjing University of Posts and Telecommunications, Nanjing 210046, China, Key Laboratory of Artificial Intelligence,  Ministry of Education, Shanghai 200240, China, and also with the Provincial Key Laboratory for Computer Information Processing Technology, Soochow University, Suzhou 215006, China (e-mail: csggao@gmail.com).
\IEEEcompsocthanksitem Juncheng Li is with the School of Communication and Information Engineering, Shanghai University, Shanghai 200444, China (e-mail: cvjunchengli@gmail.com).
\IEEEcompsocthanksitem Jian Yang is with the School of Computer Science and Technology, Nanjing University of Science and Technology, Nanjing 210094, China (e-mail: csjyang@njust.edu.cn).
\IEEEcompsocthanksitem Guo-Jun Qi is with the Research Center for Industries of the Future and the School of Engineering, Westlake University, Hangzhou 310024, China, and also with OPPO Research, Seattle, WA 98101 USA (e-mail: guojunq@gmail.com).
\IEEEcompsocthanksitem Chia-Wen Lin is with the Department of Electrical Engineering, National Tsing Hua University, Hsinchu, Taiwan 30013, R.O.C. (e-mail: cwlin@ee.nthu.edu.tw).}
}

\markboth{IEEE Transactions on Multimedia}%
{Shell \MakeLowercase{\textit{et al.}}: Bare Demo of IEEEtran.cls for IEEE Journals}

\maketitle

\begin{abstract}
Lightweight image super-resolution aims to reconstruct high-resolution images from low-resolution images using low computational costs. However, existing methods result in the loss of middle-layer features due to activation functions. To minimize the impact of intermediate feature loss on reconstruction quality, we propose a Feature Interaction Weighted Hybrid Network (FIWHN), which comprises a series of Wide-residual Distillation Interaction Block (WDIB) as the backbone. Every third WDIB forms a Feature Shuffle Weighted Group (FSWG) by applying mutual information shuffle and fusion. Moreover, to mitigate the negative effects of intermediate feature loss, we introduce Wide Residual Weighting units within WDIB. These units effectively fuse features of varying levels of detail through a Wide-residual Distillation Connection (WRDC) and a Self-Calibrating Fusion (SCF). To compensate for global feature deficiencies, we incorporate a Transformer and explore a novel architecture to combine CNN and Transformer. We show that our FIWHN achieves a favorable balance between performance and efficiency through extensive experiments on low-level and high-level tasks. Codes will be available at \url{https://github.com/IVIPLab/FIWHN}. 
\end{abstract}



\IEEEpeerreviewmaketitle

\section{Introduction}\label{introduction}
\IEEEPARstart{S}ingle-image super-resolution (SISR) has gained increasing attention due to the demand for high-resolution in various computer vision applications, including medical image analysis, security surveillance~\cite{gao2022context}, and autonomous driving. However, SISR remains a challenging problem as it involves solving an inverse problem. 
The advent of deep neural networks has significantly advanced the field of SISR. One of the pioneering works, SRCNN~\cite{dong2015image}, employed a three-layer convolutional network and outperformed traditional methods significantly. Then VDSR~\cite{kim2016accurate} increased the model depth to 20 layers and achieved better performance. Subsequently, several approaches~\cite{zhang2018image} utilized deeper networks to improve the performance of SISR. However, these methods have a high computational overhead, making them unsuitable for practical devices with limited computational resources. 

\begin{figure}[t]
	\scriptsize
	\centering
	\scalebox{0.88}{
		\begin{tabular}{lc}

            \begin{adjustbox}{valign=t}
				\begin{tabular}{c}
					\includegraphics[width=0.19\textwidth, height=0.115\textheight]{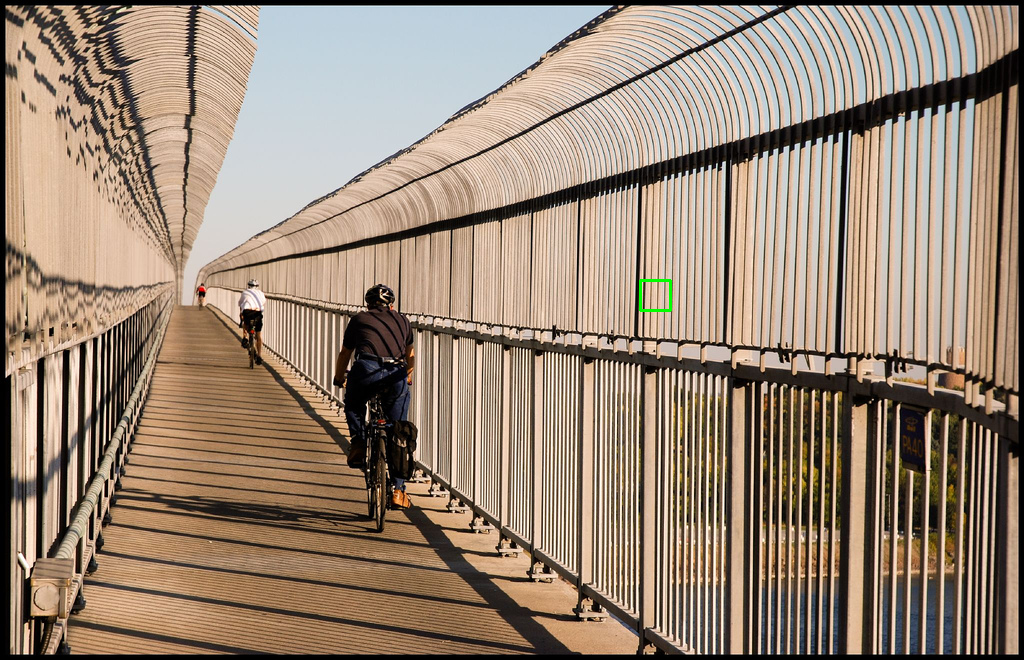} \\
					Urban100 ($\times 4$): \\
					img024 \\
				\end{tabular}
			\end{adjustbox}
			\hspace{-3mm}
			\begin{adjustbox}{valign=t}
				\begin{tabular}{ccc}
					\includegraphics[width=0.1\textwidth, height=0.045\textheight]{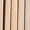} & 
					\hspace{-3mm}
					\includegraphics[width=0.1\textwidth, height=0.045\textheight]{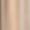} & 
					\hspace{-3mm}
					\includegraphics[width=0.1\textwidth, height=0.045\textheight]{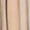} \\
					HR & \hspace{-3mm}
				  Bicubic & \hspace{-3mm}
					CT in series \\
					PSNR/SSIM & \hspace{-3mm}
					18.27/0.4629 & \hspace{-3mm}
					20.13/0.6647 \\
					\includegraphics[width=0.1\textwidth, height=0.045\textheight]{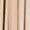} & 
					\hspace{-3mm}
					\includegraphics[width=0.1\textwidth, height=0.045\textheight]{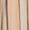} &
					\hspace{-3mm}
					\includegraphics[width=0.1\textwidth, height=0.045\textheight]{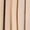} \\
				  TC in series & 
                    \hspace{-3mm}
					Parallel & \hspace{-3mm}
					\textbf{Ours} \\
					21.02/0.7024 & \hspace{-3mm}
					20.29/0.6628 & \hspace{-3mm}
					\textbf{21.27/0.7134} \\
				\end{tabular}
			\end{adjustbox}
			\\
			
	\end{tabular} }
	\caption{Comparisons of different interaction schemes between the CNN and Transformer. ``CT in series'' is shown in Figure~\ref{combine} (a), denoting the series connection of CNN with Transformer; ``TC in series'' is shown in Figure~\ref{combine} (b), denoting the series connection of Transformer with CNN; ``Parallel'' is shown in Figure~\ref{combine} (c), denoting the parallel connection between CNN and Transformer; ``Ours'' is shown in Figure~\ref{combine} (d), denoting the potential interaction between CNN and Transformer.}
    \vspace{-3mm}
	\label{cnn-trans-position}
\end{figure}

To reduce the model size, many existing approaches have focused on designing efficient model structures,  which include weight sharing~\cite{gao2022feature}, multi-scale structures~\cite{zhang2021pffn}, strategies for neural structure search~\cite{chu2021fast}, grouped convolution~\cite{ahn2018fast}. However, existing approaches often overlook the loss of intermediate information caused by activation functions like ReLU. This issue has been demonstrated by MobileNetV2~\cite{sandler2018mobilenetv2}, where the reduction in intermediate information during increases in network depth can negatively affect the quality of image reconstruction. We propose the Feature Interaction Weighted Hybrid Network (FIWHN) to address this concern while maintaining lightweight models. Specifically, our Convolutional Neural Network (CNN) part incorporates wide-residual attention-weighted units, which consist of Wide Identical Residual Weight (WIRW) and Wide Convolutional Residual Weighting (WCRW). These units help compensate for the lost intermediate features by obtaining a broader feature map before applying the activation function. Additionally, we adopt Wide-residual Distillation Interaction Blocks (WDIB) with a lattice structure~\cite{luo2022lattice}. The WDIB includes two paired skip connections and adaptive combinations of wide residual blocks that utilize attention-based connection weights. As a result, we achieve a compact network with strong expressive power. The Wide-Residual Distillation Connection (WRDC) framework and the Self-Calibrating Fusion (SCF) unit facilitate the distillation and fusion of split features from different classes within the WDIB, thereby enhancing its generalization capability. Multiple WDIBs are combined to form a Feature Shuffle Weighted Group (FSWG), which leverages information from the middle layers at the group level through blending, fusion, and weighting of the output features from each WDIB. 


Recently, Transformer-based methods have demonstrated great potential for SISR. For instance, SwinIR-light~\cite{liang2021swinir} leveraged the Transformer's ability to model long-range dependencies by employing a sliding window mechanism to address the issue of uncorrelated edges between image patches. Hybrid networks~\cite{lu2021efficient,gao2022lightweight,zhang2022efficient,li2023cross} that combine CNN and Transformer have also exhibited advantages over pure CNN or pure Transformer. Therefore, in our method, we also introduce the Transformer to enable effective long-range modeling. By combining CNN and Transformer, the weights of these models can be adjusted based on the information extracted from each other during the training process. However, existing methods, represented in Figure~\ref{combine} (a) and (b), often struggle to effectively integrate local and global information flow, resulting in the generation of ambiguous artifacts, as shown in Figure~\ref{cnn-trans-position}. To address this limitation, we propose an improved approach that combines CNN and Transformer. Our goal is to facilitate a stronger interaction between the features extracted by both networks, thus enhancing the overall performance.

In summary, the main contributions are listed as follows:
\begin{itemize}
  \item We propose wide-residual weighting units for SISR, which consist of WIRW and WCRW. They effectively mitigate the negative impact of intermediate feature loss by incorporating a wide residual mechanism.
  \item We introduce WRDC, which enhances information flow by leapfrogging features at different levels within the WDIB. Additionally, we propose a SCF, allowing for a more precise and adaptive feature combination.
  \item We present a novel interaction framework that enhances communication between different levels of features. This includes the FSWG in the CNN part and an interaction framework between CNN and Transformers.
\end{itemize}

In this work, we have expanded on the following aspects compared to our conference version paper~\cite{gao2022feature}:
\begin{itemize}
  \item To address the issue of missing global features, we incorporate the Transformer component by designing a novel combinatorial framework that promotes better information flow, which outperforms existing frameworks.
  \item We conduct comprehensive experiments covering both real-world as well as segmentation scenarios, and the results consistently support the competitiveness of our proposed method in the SISR domain.
\end{itemize}

\begin{figure*}
\begin{center}
\includegraphics[width=0.9\linewidth]{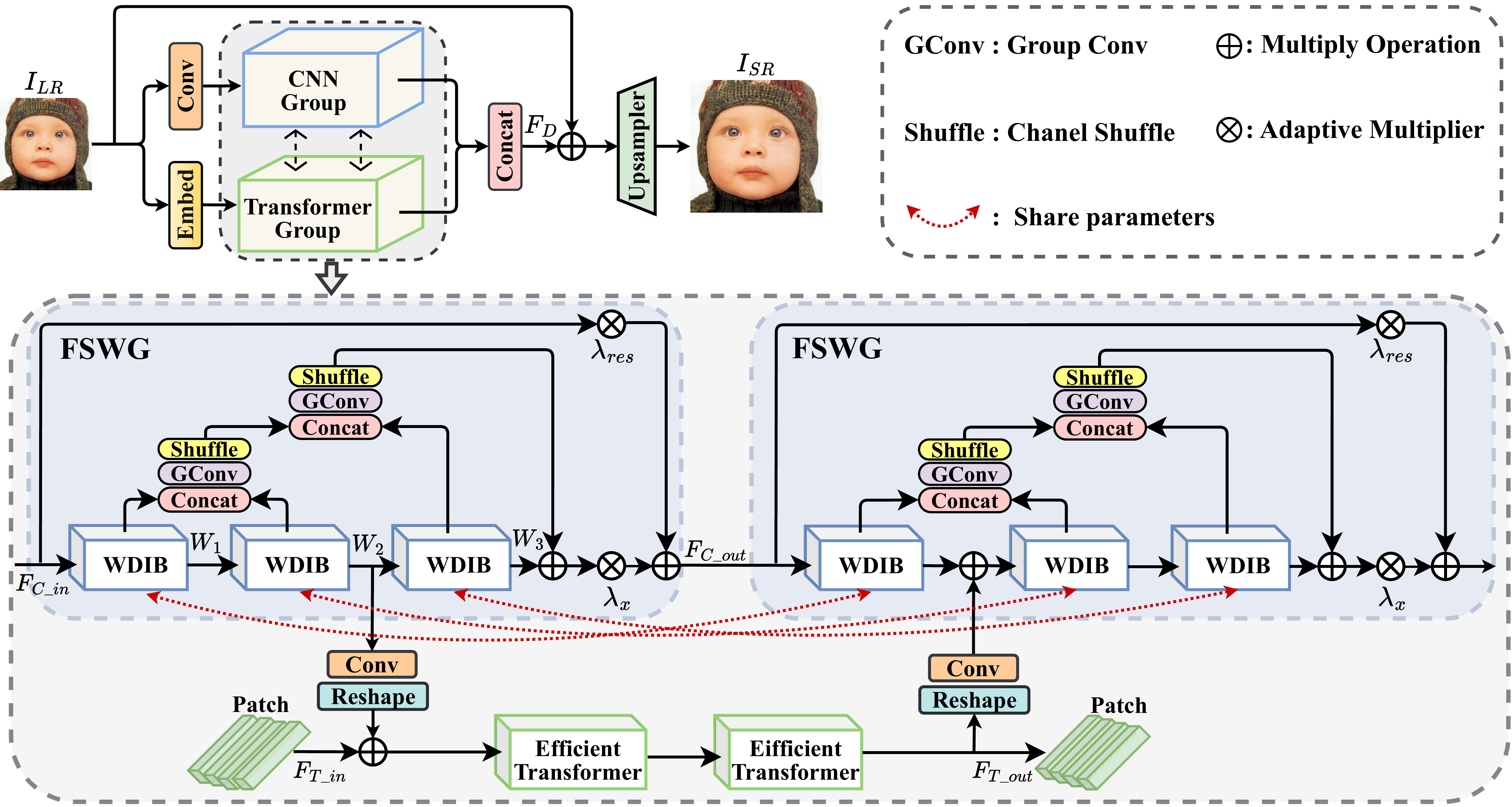}
\end{center}
\caption{Architecture of our proposed Feature Interaction Weighted Hybrid Network (FIWHN).}
\label{FIWHN}
\vspace{-0.1cm}
\end{figure*}

\section{RELATED WORK}
In this study, we specifically focus on lightweight SISR models, wide-residual weighting learning, and Transformer-based SISR models.

\subsection{Lightweight SISR Models}
To make SISR applicable to real-world applications with limited computational resources, there has been significant research focused on developing lightweight SISR models. Existing research in this area can be broadly categorized into three main approaches: efficient model structure design-based methods~\cite{hui2019lightweight, luo2022lattice, gao2022feature}, pruning or quantification techniques-based methods~\cite{li2020pams}, and knowledge distillation-based methods~\cite{lee2020learning}. Efficient model structure design primarily involves designing model architectures specifically tailored for lightweight SISR. For instance, some models adopt recursive cascading to learn feature representations across different layers~\cite{ahn2018fast}, while others reuse intermediate layer features through recursive learning~\cite{gao2022feature}. CFSR~\cite{wu2024transforming} utilized large kernel convolutions to capture long-range features. Strategies like channel splitting and hierarchical distillation have also been explored in models like IMDN~\cite{hui2019lightweight} to extract features at different levels. Additionally, applying neural architecture search (NAS) in SISR, as demonstrated by FALSR~\cite{chu2021fast}, has introduced new possibilities for structural design-based methods. Knowledge distillation methods leverage the knowledge transfer from pre-trained large teacher models to smaller student models to improve their performance~\cite{lee2020learning}. Pruning or quantization techniques aim to reduce the model size without sacrificing accuracy~\cite{li2020pams}. Our FIWHN belongs to the category of efficient structure design-based method, which investigates the network structure in terms of inter-block design, aiming to efficiently combine our units to achieve a lightweight model.

\subsection{Wide-Residual Weighting Learning}
Numerous studies~\cite{kim2016accurate, zhang2018image} have investigated the relationship between network depth and performance in deep learning models. Initially, it was believed that deeper networks would lead to better performance.  For instance, VDSR~\cite{kim2016accurate} utilized a 20-layer network, and RCAN~\cite{zhang2018image} went even further with a network depth exceeding 800 layers. However, subsequent research revealed that increasing network depth does not necessarily result in better performance and may lead to a decline. Studies, such as MobileNetV2~\cite{sandler2018mobilenetv2}, have shown that the extensive use of activation functions in models can contribute to feature degradation. The widely-used ReLU function, in particular, can cause some neurons to become ``dead'', resulting in the loss of intermediate features. This degradation becomes more pronounced as the network grows deeper. Building upon the findings of WDSR~\cite{yu2018wide}, which demonstrated that models with wider features before the Relu activation layer can achieve better performance, we introduce adaptive multipliers that allow for the adaptive adjustment of wide-residual weighting units during training. Consequently, our proposed residual unit enables efficient feature extraction while maintaining a lightweight structure.

\subsection{Transformer-based SISR Models}
Transformers have recently demonstrated their remarkable capabilities in computer vision. As a result, Transformer-based approaches for SISR have gained significant attention. SwinIR-light~\cite{liang2021swinir} initially demonstrated state-of-the-art performance by introducing Transformer-based strategies to the SISR task. Building on this advancement, ESRT~\cite{lu2021efficient} and LBNet~\cite{gao2022lightweight} combined lightweight CNN with lightweight Transformers in the SISR task, achieving a good balance across multiple evaluation metrics. ELAN~\cite{zhang2022efficient} accelerates models by grouping multi-scale self-attentive schemes and attention-sharing mechanisms. SCTANet ~\cite{bao2023sctanet} and ELSFace~\cite{qi2023efficient} both proposed to improve performance by utilizing a parallel structure involving both CNNs and transformers. CFIN~\cite{li2023cross} and NGSwin~\cite{choi2023n} further enhanced performance by introducing the context to expand the perceptual field. However, unlike these previous works, we explore a more efficient structure of combining CNNs with Transformers to further improve the interaction ability of local and global features.

\begin{figure*}
\begin{center}
\includegraphics[width=0.95\linewidth]{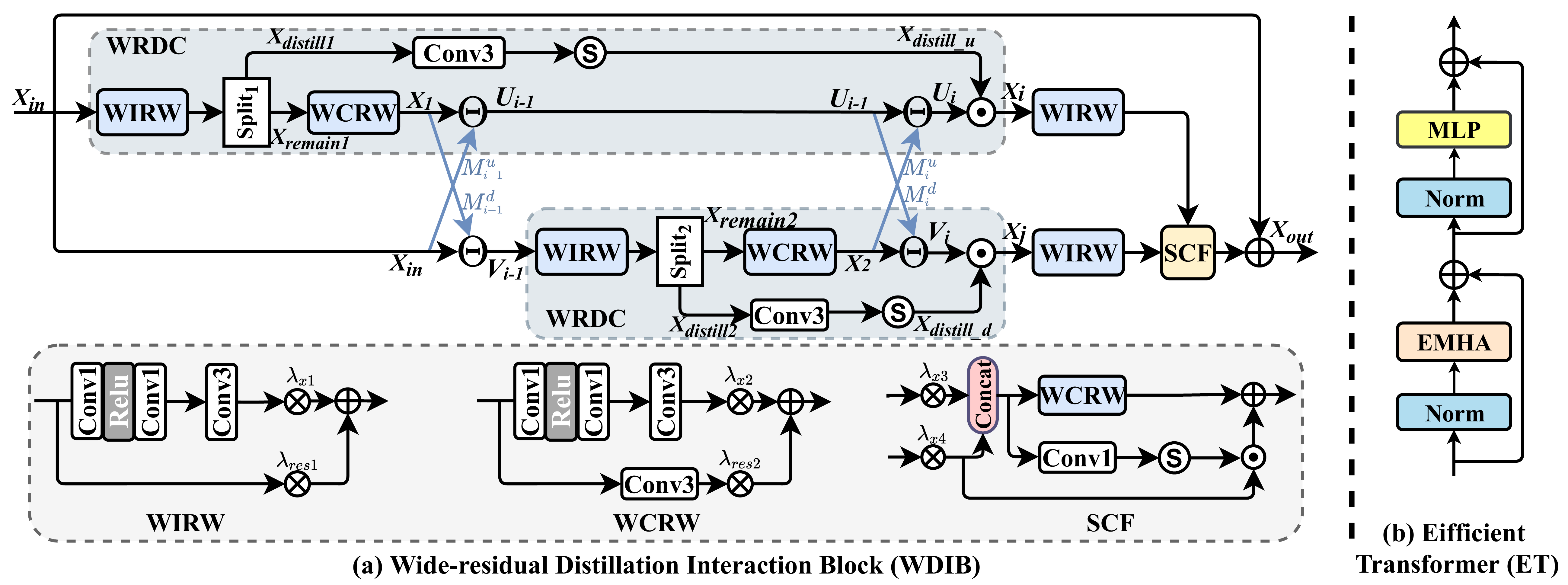}
\end{center}
\caption{(a) Structure of the Wide-residual Distillation Interaction Block (WDIB). The $M_i$ and $M_{i-1}$ represent the combination coefficient learning, which can be understood in Figure~\ref{PC}, and $ \odot $ represents the operation of multiplication, \textcircled{s} represents the sigmoid function, $\Theta ({x_i},{y_i}) = {x_i} + {y_i}{M_i}({y_i})$; (b) The structure of the Efficient Transformer (ET).}
\label{WDIB}
\vspace{-0.2cm}
\end{figure*}

\section{PROPOSED METHOD}

In this section, we present an overview of FIWHN. First, we outline the general structure, which includes the backbone of the CNN and Transformer and their interaction. Next, we introduce our proposed WDIB. Finally, we provide details about the supervision functions used in training the model.

\subsection{Feature Interaction Weighting Hybrid Network}
\textbf{Overview.} As presented in Figure~\ref{FIWHN}, FIWHN consists of three main components: the dimensional transformation part, the interaction part between CNN and Transformer, and the upsampling part. In this setup, ${I_{LR}}$ and ${I_{SR}}$ represent the input LR image and the SR image, respectively.

First, the dimensional transformation part consists of a convolutional layer ${G_{Conv}}$ and an embedding module ${G_{Embed}}$. The process can be expressed as
\begin{equation}
{F_C} = {G_{Conv}}({I_{LR}}),
{F_T} = {G_{Embed}}({I_{LR}}).
\end{equation}
Then, the respective outputs ${F_C}$ and ${F_T}$ are sent to the CNN and Transformer interaction stage:
\begin{equation}
{F_D} = {G_{Concat}}(CNN({F_C}) \leftrightarrow Trans({F_T})),
\end{equation}
where $CNN$ refers to the CNN group, $Trans$ denotes the Transformer group, $ \leftrightarrow $ represents the information exchange process between CNN and Transformer, and ${G_{Concat}}$ denotes the concat operation. The depth feature ${F_D}$ is then passed to the upsampling part ${G_{Upsample}}$ along with the input to complete the image reconstruction:
\begin{equation}
{I_{SR}} = {G_{Upsample}}({I_{LR}} + {F_D}).
\end{equation}

\textbf{Feature Shuffle Weighted Group (FSWG).} In FSWG, we have incorporated a feature shuffling and fusion mechanism to effectively combine, group, and shuffle features from different receiver domains. As depicted in Figure~\ref{FIWHN}, FSWG serves as the backbone component of the CNN part and consists of 3 interacting WDIBs. Specifically, we progressively blend and shuffle the output features of adjacent WDIB. This cascade operation, denoted as ${G_{CGS}}$, can be represented as:
\begin{equation}
{G_{CGS}} = {G_{Shuffle}}({G_{GConv}}({G_{Concat}}[{x_i},{x_{i + 1}}])),
\end{equation}
where ${G_{GConv}}$ represents the operation of group convolution, ${G_{Shuffle}}$ represents the operation of channel shuffle like ShuffleNet~\cite{zhang2018shufflenet}, ${x_i}$ and ${x_{i + 1}}$ represent the output features of the two blocks to be fused,  respectively. Additionally, adaptive multipliers are applied to both the fused features between blocks and the original input features. This allows the module to adjust its weights dynamically in addition to the weight updates during training. Let's define the input as ${F_{C\_in}}$ and the output as ${F_{C\_out}}$. The process is:
\begin{equation}
{F_{CGS}} = {G^2}_{CGS}({G^1}_{CGS}({W_1},{W_2}),{W_3}),
\end{equation}
\begin{equation}
{F_{C\_out}} = {\lambda _x}({F_{CGS}} + {W_3}) + {\lambda _{res}}{F_{C\_in}},
\end{equation}
where ${W_i}$ represents the output of the $i$-th WDIBs, ${G^i}_{CGS}$ denotes the function of the $i$-th ${G}_{CGS}$, ${F_{CGS}}$ represents the output features obtained from different blocks after a series of fusion, grouping, and shuffling operations, ${\lambda _{x}}$ and ${\lambda _{res}}$ are adaptive multipliers, which can be automatically learned and modified during the training process to optimize the model. 

\begin{figure}[t]
\centering
\includegraphics[width=0.45\textwidth,trim=0 0 20 0]{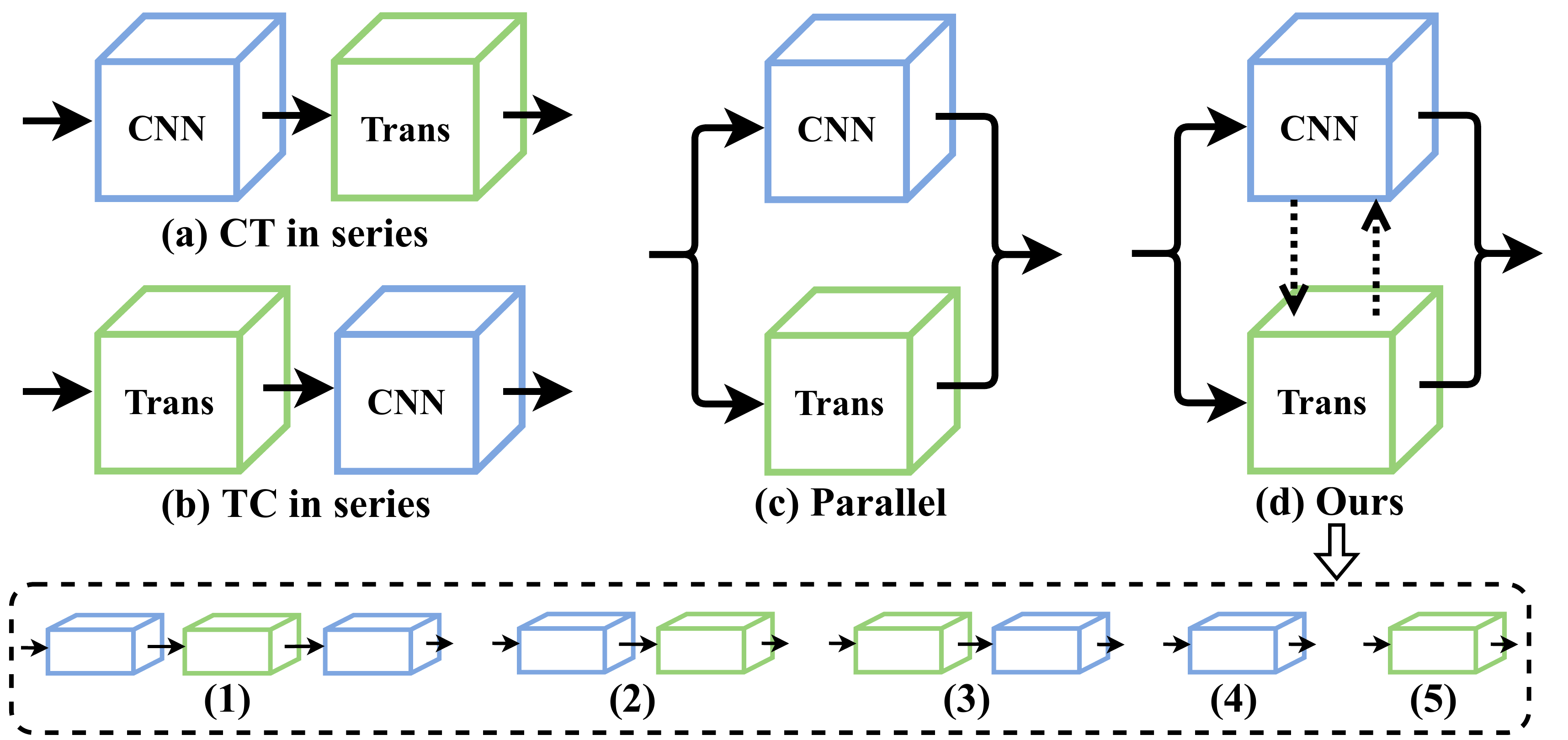}
\caption{Exploring how to combine CNN and Transformer efficiently and the potential of our method for multiple combinations of both.}
\label{combine}
\end{figure}

\textbf{The Interaction of CNN and Transformer.} The integration architectures can be categorized into three main types, as depicted in Figure~\ref{combine} and labeled as (a), (b), and (c). Methods such as ESRT~\cite{lu2021efficient}, LBNet~\cite{gao2022lightweight}, and CFIN~\cite{li2023cross} adopt the structures depicted in (a) and (b) in Figure~\ref{combine}. These methods concatenate CNN and Transformer modules to focus on local and global features in a batch-wise manner. Another set of methods, such as Faceformer~\cite{wang2022faceformer}, utilize the structure of (c) in Figure~\ref{combine}. They connect the CNN and Transformer modules in parallel and then fuse the extracted local features with the global features, enabling feature extraction for image reconstruction. However, these methods overlook the significance of interactions between local features, extracted from the middle layer of the model, and global features. Our proposed interaction approach addresses this concern by facilitating the free flow of local and global patterns within the network, enabling them to mutually guide each other. This interaction approach enables multiple interactions between local and global features, as illustrated in Figure~\ref{combine} (d). Specifically, as demonstrated in Figure~\ref{FIWHN}, the local features ${W_2}$ from the first FSWG are combined with the global input features ${F_{T\_in}}$ from the Transformer branch after undergoing dimension and shape transformation through ${G_{CR}}$. This fusion of features is then globally modeled using Efficient Transformer ${G_{ET}}$. The resulting features, denoted as ${F_{T\_out}}$, are utilized in the subsequent stages of global modeling and also fed into the second FSWG for local feature extraction. The process can be expressed as follows:
\begin{equation}
{F_{T\_out}} = {G_{ET}}({G_{ET}}({G_{CR}}({W_2}) + {F_{T\_in}})).
\end{equation}

\textbf{Efficient Transformer (ET).} In the context of lightweight models with limited network depth, using a pure CNN model alone is insufficient for reconstructing high-quality images. To address this issue, we propose a solution that involves compressing the size of the CNN and incorporating an efficient Transformer module to capture long-distance dependencies in images. Specifically, as shown in Figure\ref{WDIB} (b), we adopt the design philosophy of ESRT\cite{lu2021efficient} for multi-headed attention (MHA), which consumes less GPU training memory. This is achieved by splitting the generated tokens $Q$, $K$, and $V$ from the linear layer along the width and height dimensions. The formulation of this process is as follows:
\begin{equation}
\begin{array}{l}
({Q_1}...{Q_n}),{\rm{ }}({K_1}...{K_n}),{\rm{ }}({V_1}...{V_n}){\rm{ }} = {\rm{ }}Split(Q,{\rm{ }}K,{\rm{ }}V).
\end{array}
\end{equation}
Then, the sub-token obtained after subsequent splitting is matrix multiplied on a field of only $\frac{1}{n}$ (where $n$ represents the number of feature splits, and we chose four splits) of the original perception. This effectively reduces memory consumption. Finally, the sub-attention obtained from the matrix dot product is merged to obtain the final self-attention. The entire process can be described as follows:
\begin{equation}
\begin{array}{l}
{O_i} = Attentio{n_i}({Q_i},{K_i},{V_i}) = {\rm{softmax}}(\frac{{{Q_i}K_i^T}}{{\sqrt {{d_k}} }}){{\rm{V}}_i},\\
\end{array}
\end{equation}
\begin{equation}
\begin{array}{l}
Attention(Q,K,V){\rm{ }} = {\rm{ }}Concat({O_1},...,{O_n}).\\
\end{array}
\end{equation}

\begin{figure}[t]
\centering
\includegraphics[width=0.46\textwidth,trim=0 0 20 0]{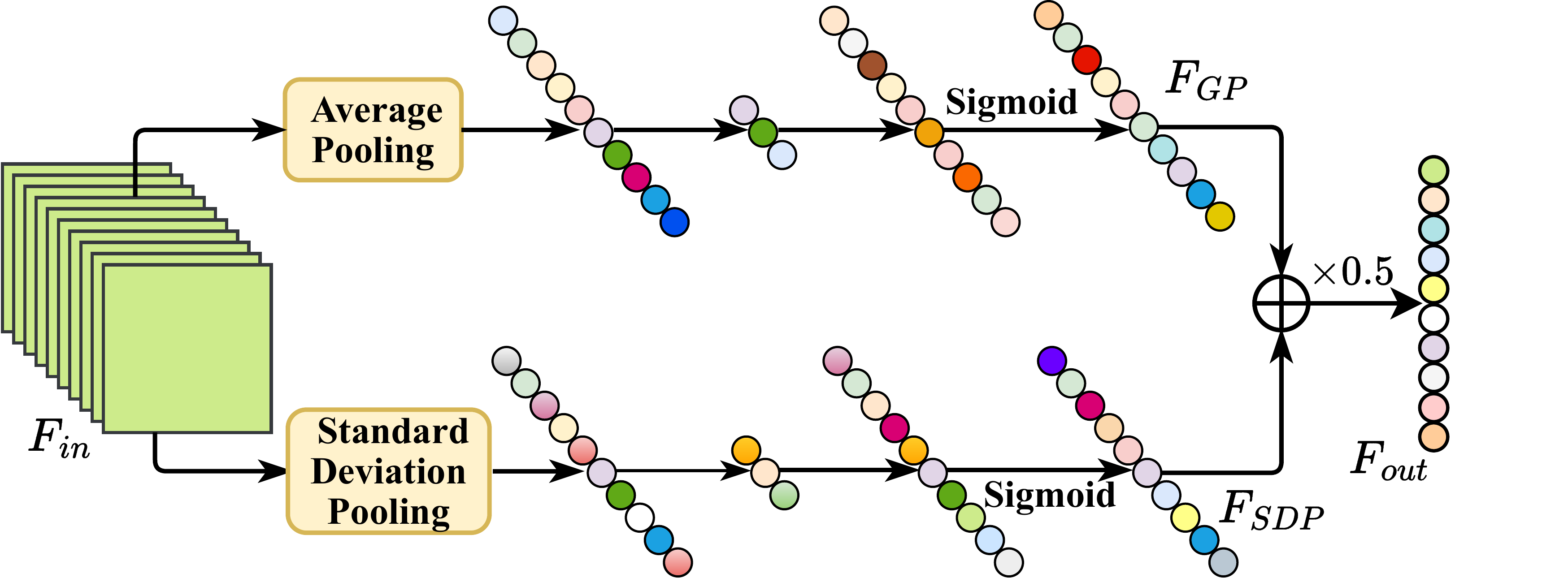}
\caption{Details of the combination coefficient learning, which corresponds to $M_i$ and $M_{i-1}$ in Figure~\ref{WDIB} (a).}
\label{PC}
\end{figure}

\subsection{Wide-Residual Distillation Interaction Block}

\textbf{Lattice structure.} Inspired by the advantages of lattice blocks~\cite{luo2022lattice}, as depicted in Figure~\ref{WDIB} (a), we employ this structure to combine wide-residual weighted units. The structure consists of two paired skip connections designed to connect upper and lower features by combining learning coefficients. Each paired skip connection introduces a distinct combination pattern for the residual units. Additionally, we incorporate feature splitting and information refinement when combining residual units. We also leverage the concept of feature distillation~\cite{hui2019lightweight} to efficiently perform feature selection and fusion. Since too many distillations may affect the efficiency of the model, we only distill the information once in each of the upper and lower branches to ensure enhanced generalization of the model. Specifically, for the input feature ${X_{in}}$ that feeds into the upper and lower branches, we define ${F_{ir}}$ as the WIRW unit and ${F_{cr}}$ as the WCRW unit. The operation of the upper branch can be described as:
\begin{equation}
\begin{array}{l}
{X_{remain1}},{X_{distill1}} = Split({F_{ir}}({X_{in}})),\\
\end{array}
\end{equation}
\begin{equation}
\begin{array}{l}
{X_1} = {F_{cr}}({X_{remain1}}).\\
\end{array}
\end{equation}
The upper and lower branches are then connected via the first paired skip connections. This process can be described as:
\begin{equation}
\begin{array}{l}
{V_{i - 1}} = \Theta ({X_{in}},{X_1}) = {X_{in}} + {X_1}M_{_{i - 1}}^u({X_1}),\\
\end{array}
\end{equation}
\begin{equation}
\begin{array}{l}
{U_{i - 1}} = \Theta ({X_1},{X_{in}}) = {X_1} + {X_{in}}M_{_{i - 1}}^d({X_{in}}),\\
\end{array}
\end{equation}
where $M_{_{i - 1}}^u$ and $M_{_{i - 1}}^d$ represent the two combined coefficient learning mechanisms that connect the upper and lower branches in the first paired skip connections, respectively. As shown in Figure~\ref{PC}, when the input is ${F_{in}}$, the output ${F_{out}}$ of the mechanism can be formulated as:
\begin{equation}
\begin{array}{l}
{F_{GP}} = Sigmoid\left( {Con{v_ \uparrow }\left( {Con{v_ \downarrow }\left( {Avg\left( {{F_{in}}} \right)} \right)} \right)} \right),\\
\end{array}
\end{equation}
\begin{equation}
\begin{array}{l}
{F_{SDP}} = Sigmoid\left( {Con{v_ \uparrow }\left( {Con{v_ \downarrow }\left( {Std\left( {{F_{in}}} \right)} \right)} \right)} \right),\\
\end{array}
\end{equation}
\begin{equation}
\begin{array}{l}
{F_{out}} = ({F_{GP}} + {F_{SDP}})/2,\\
\end{array}
\end{equation}
where ${Avg}$ and ${Std}$ are average pooling and standard deviation pooling, respectively, ${Con{v_ \downarrow }}$ and ${Con{v_ \uparrow }}$ are convolution operations for reducing channel dimensions and ascending channel dimensions, respectively, and ${F_{GP}}$ and ${F_{SDP}}$ are the outputs of the upper and lower branches, respectively. Compared to the average pooling used in traditional channel attention, we have incorporated a standard difference pooling branch here for improved visualization, as verified in~\cite{hui2019lightweight}. ${U_{i - 1}}$ and ${V_{i - 1}}$ represent the output features of the upper and lower branches, respectively, after passing through the first paired skip connections. Subsequently, ${U_{i-1}}$ and ${V_{i-1}}$ are then fed into the second paired skip connections, which follow a similar process as the first paired skip connections. It can be listed as:

\begin{equation}
\begin{array}{l}
{X_{remain2}},{X_{distill2}} = Split({F_{ir}}({V_{i - 1}})),\\
\end{array}
\end{equation}
\begin{equation}
\begin{array}{l}
{X_2} = {F_{cr}}({X_{remain2}}),\\
\end{array}
\end{equation}
\begin{equation}
\begin{array}{l}
{V_i} = \Theta ({X_2},{U_{i - 1}}) = {X_2} + {U_{i - 1}}M_i^u({U_{i - 1}}),\\
\end{array}
\end{equation}
\begin{equation}
\begin{array}{l}
{U_i} = \Theta ({U_{i - 1}},{X_2}) = {U_{i - 1}} + {X_2}M_{_i}^d({X_2}).\\
\end{array}
\end{equation}

Similar to before, ${U_i}$ and ${V_i}$ represent the output features of the upper and lower branches after the second skip connection, respectively. $M_i^u$ and $M_i^d$ denote the combined coefficient learning mechanisms above and below the connection of the second paired skip connection. Simultaneously, the features extracted from the upper and lower branches undergo a nonlinear transformation through the operation of convolution followed by sigmoid activation. This process completes the non-linearization of the coarse features. As a result, we obtain the coarse features ${X_{distill\_u}}$ from the upper branch and the coarse features ${X_{distill\_d}}$ from the lower branch. The process can be expressed as:
\begin{equation}
\begin{array}{l}
{X_{distill\_u}} = {F_{sigmoid}}({F_{conv3}}({X_{distill1}})),\\
\end{array}
\end{equation}
\begin{equation}
\begin{array}{l}
{X_{distill\_d}} = {F_{sigmoid}}({F_{conv3}}({X_{distill2}})).\\
\end{array}
\end{equation}

Next, obtained coarse features ${X_{distill\_u}}$ and ${X_{distill\_d}}$ interact with the fine features ${U_i}$ and ${V_i}$. They are modulated by the attention-based combined coefficient learning mechanism to achieve feature blending with varying degrees of refinement. Finally, the blended features ${X_i}$ and ${X_j}$ are fused using our well-designed Self-Calibration Fusion (SCF) module to facilitate the adaptive fusion of the blended features obtained from the two branches. Original input features are retained through residual concatenation. This can be formulated as:
\begin{equation}
\begin{array}{l}
{X_{out}} = {F_{SCF}}({F_{ir}}({X_i}),{F_{ir}}({X_j})) + {X_{in}},\\
\end{array}
\end{equation}
where ${F_{SCF}}$ represents the SCF module. Within the SCF module, the outputs of the upper and lower branches perform weighted connections. Subsequently, different levels of refinement are applied to the fused features, resulting in a diverse range of information being incorporated into the final fused features. The module adjusts its weights during training, leading to improved performance compared to standard fusion techniques.

\begin{table}[t]
	\centering
	\small
	\caption{Analysis of the effect of the wide residual mechanism on WIRW and WCRW.}
	\label{tab:RB}
	\scalebox{0.88}{
	\begin{tabular}{|c|c|c|c|c|}
        \hline
		\multirow{2}{*}{Methods} & \multirow{2}{*}{Channels}  & \multirow{2}{*}{Params} & \multirow{2}{*}{Multi-adds} & Set5($\times 4$)   \\
		\cline{5-5}
		& & & & PSNR / Time  \\
		\hline
		\hline
        Baseline  & 32    & 223K  & 12.69G     & 31.75 / 7.22ms          \\
        FIWHN     & 64  & 147K  & 4.46G     & 31.76 / \textbf{5.72ms}           \\
        FIWHN     &120    &175K  & 9.89G & \textbf{31.83} / 6.49ms  \\ 
		\hline
	\end{tabular}
	}
\end{table}

\begin{table}
	\centering
	\caption{Impact analysis of different module combinations in the WDIB framework, where $\otimes$ is the adaptive multiplier.}
	\label{tab:WRDC-SCF}
	\scalebox{0.83}{
	\begin{tabular}{|c|cccc|c|c|c|}
        \hline
		\multirow{2}{*}{Methods} & \multirow{2}{*}{WRDC}  & \multirow{2}{*}{SCF} & \multirow{2}{*}{BI} &\multirow{2}{*}{$\otimes$} & \multirow{2}{*}{Params}  & \multirow{2}{*}{Multi-adds}  & Set5($\times 4$)   \\
		\cline{8-8}
		& & & & & & & PSNR / SSIM  \\
		\hline
		\hline
        Baseline  &\XSolid   &\XSolid  &\XSolid  &\XSolid & 59.3K & 3.36G  & 31.17 / 0.8791    \\
		Case 1  &\Checkmark      & &  & & 49.2K & 2.11G  & 31.17 / 0.8799  \\
        Case 2   &       &\Checkmark &  & & 69.5K & 3.62G  & 31.35 / 0.8824 \\
        Case 3   &       & &\Checkmark  & & 59.3K & 3.36G  & 31.22 / 0.8791\\
		Case 4  &    &  &  &\Checkmark & 59.3K & 3.36G  & 31.21 / 0.8794  \\
        FIWHN &\Checkmark     &\Checkmark  &\Checkmark &\Checkmark   & 69.9K & 2.57G   &{\bf31.42 / 0.8835}  \\
		\hline
	\end{tabular}
	}
\end{table}

\textbf{Wide-Residual Distillation Connection (WRDC).}
As depicted in Figure~\ref{WDIB}, the Wide-Residual Distillation Connection (WRDC) is a key component of the model, comprising Wide Convolutional Residual Weighting (WCRW) and Wide Identical Residual Weighting (WIRW) units, as well as residual connections for feature refinement. Both WIRW and WCRW introduce a wide range of activation mechanisms to mitigate the loss of intermediate layer features and extract more expressive features. For WIRW, the wide residual mechanism splits the original residual's initial $3 \times 3$ convolution into two $1 \times 1$ convolution, which the first $1 \times 1$ convolution increases the channel dimension substantially to handle the subsequent activation functions and reduce feature loss, while the second $1 \times 1$ convolution is then used for channel dimensionality reduction, preventing an excessive number of parameters that would arise from using $3 \times 3$ convolutional layers for feature extraction. For an input feature $x$, the broad features obtained through this process can be expressed as:
\begin{equation}
\begin{array}{l}
{x_{wide}} = {F_{conv3}}({F_{conv1 \downarrow }}({F_{relu}}({F_{conv1 \uparrow }}(x)))),\\
\end{array}
\end{equation}
where ${F_{conv1 \uparrow }}$ represents the channel up-dimensioning operation of the first $1 \times 1$ convolution, ${F_{conv1 \downarrow }}$ represents the channel down-dimensioning operation of the second $1 \times 1$ convolution, ${F_{relu}}$ denotes the Relu function used for non-linearization, and ${F_{conv3}}$ refers to the $3 \times 3$ convolution. Since all the high-dimensional channel operations are performed on the $1 \times 1$ convolution, they do not impose a significant computational burden. Subsequently, adaptive multipliers are incorporated into both the main branch and the residual branch of the residual block, enabling autonomous adjustment of the weights during training. It is worth noting that WCRW has additional $3 \times 3$ convolution layers added to its shortcut path compared to WIRW. This ensures a match with the original input channel size after channel splitting. Consequently, the outputs ${y_{wirw}}$ and ${y_{wcrw}}$ for WIRW and WCRW can be respectively expressed as:
\begin{equation}
\begin{array}{l}
{y_{wirw}} = {\lambda _{x1}}{x_{wide}} + {\lambda _{res1}}x,\\
\end{array}
\end{equation}
\begin{equation}
\begin{array}{l}
{y_{wcrw}} = {\lambda _{x2}}{x_{wide}} + {\lambda _{res2}}{F_{conv3}}(x),\\
\end{array}
\end{equation}
where ${\lambda _{xk}}$ and ${\lambda _{resk}}$ ($k$=1,2) represent the adaptive weighted multipliers of the $k$-th wide residual weighted unit. Additionally, a convolutional layer is introduced in the distillation connection section to expand the dimensionality of the split channels. The obtained coarse features are then subjected to Sigmoid function nonlinearities, resulting in low-frequency feature maps. Finally, these features are multiplied with the high-frequency feature maps obtained through the combined action of wide residual units and combined coefficient learning to achieve the interaction of various types of pattern features.

\begin{table}
	\centering
	\small
	\caption{Evaluate the effectiveness of our WDIB.}
	\label{tab:WDIB-Compare}
	\scalebox{0.95}{
	\begin{tabular}{|c|c|c|c|c|}
        \hline
		\multirow{2}{*}{Methods} & \multirow{2}{*}{Depth}  & \multirow{2}{*}{Params} & \multirow{2}{*}{Multi-adds} & Set5($\times 4$)   \\
		\cline{5-5}
		& & & & PSNR / Time  \\
		\hline
		\hline
        IMDB~\cite{hui2019lightweight}   & 32    & 60.0K  & 3.42G     &  31.40 / 4.37ms       \\
        RFDB~\cite{liu2020residual}      & 24    & 63.2K  & 3.51G     &  31.36 / 3.51ms \\ 
        LB~\cite{luo2022lattice}      & 39   & 65.2K   & 3.66G     & 31.34 / 4.41ms  \\ 
        HPB~\cite{lu2021efficient}       & 48   & 64.5K   & 3.78G     & 31.36 / 6.81ms  \\ 
        LFFM~\cite{gao2022lightweight}   & 25   & 61.2K   & 3.47G     &  31.37 / 3.05ms \\ 
        \textbf{WDIB}   & 26   & 61.0K   & 2.49G     & \textbf{31.44} / \textbf{3.02ms}  \\ 
		\hline
	\end{tabular}
	}
\end{table}

\begin{table}[t]
	\centering
	\small
	\caption{Experiments about the performance of various combined CNN and Transformer architectures setting.}
	\scalebox{0.95}{
	\begin{tabular}{|c|c|c|c|c|c|}
		\hline
		{Scale} & {Architecture}  & {Set14} & B100  & Urban100 & {Manga109}\\
		\hline
		\hline
		\multirow{4}{*}{$\times 4$} &  CT in series  & 28.72  & 27.63  & 26.40  & 30.89\\
		&  TC in series  & 28.71  & 27.64  & 26.33  & 30.75\\
		&  Parallel  & 28.61  & 27.58  & 26.18  & 30.56\\
        &  Ours  &\textbf{28.76}    &\textbf{27.68}  & \textbf{26.57} &\textbf{30.93}\\
        \hline
	\end{tabular}
	}
	\label{tab:combine}
\end{table}

\subsection{Loss Function}
For the pairs $\left\{ {I_{LR}^i, I_{HR}^i} \right\}_{i = 1}^N$ in the training set, the reconstruction loss of our method FIWHN during training can be expressed as:
\begin{equation}
Loss(\theta ) = \mathop {{\rm{argmin}}}\limits_\theta  \frac{1}{N}\sum\limits_{i = 1}^N {{{\left\| {FIWHN(I_{LR}^i) - I_{HR}^i} \right\|}_1}},
\end{equation}
where $N$ represents the number of LR-HR pairs in the training set, and $\theta$ represents the parameter size of FIWHN.

\begin{figure*}[t]
\centering 
\begin{overpic}[width=0.99\linewidth]{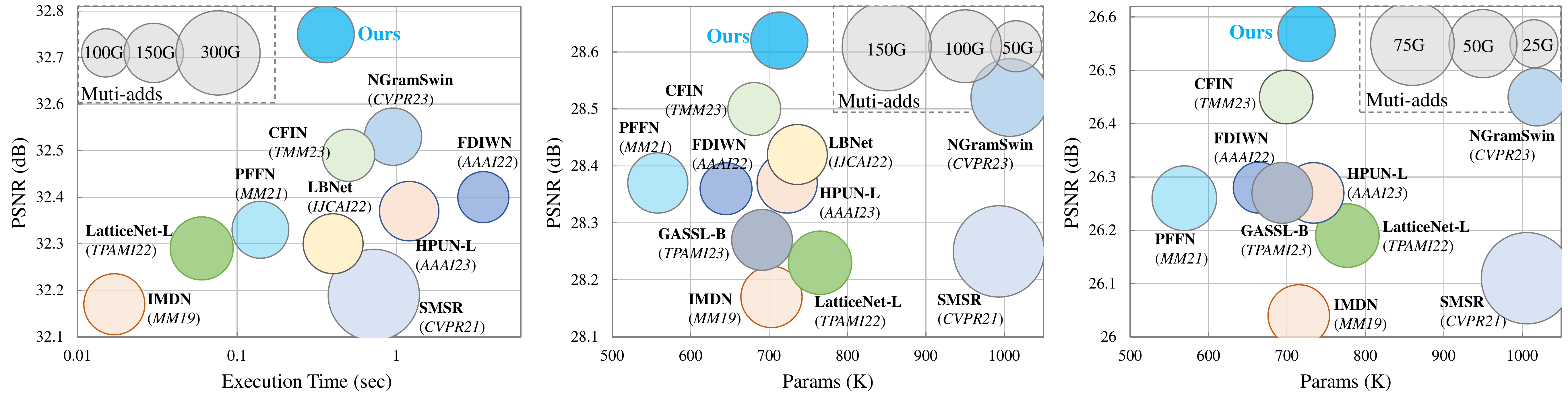}
\put(2,-2){\color{black}{\fontsize{8pt}{1pt}\selectfont (a) PSNR, Muti-adds, and Speed Tradeoffs ($\times 2$).}}
\put(35,-2){\color{black}{\fontsize{8pt}{1pt}\selectfont (b) PSNR, Parmas and Muti-adds Tradeoffs ($\times 3$).}}
\put(69,-2){\color{black}{\fontsize{8pt}{1pt}\selectfont (c) PSNR, Parmas and Muti-adds Tradeoffs ($\times 4$).}}
\end{overpic}
\vspace{0.5cm}
\caption{Model complexity analysis at each scale on the Urban100 test set.}
\label{time_compare}
\vspace{-0.2cm}
\end{figure*}

\begin{figure}[t]
\centering
\includegraphics[width=0.45\textwidth,trim=0 0 20 0]{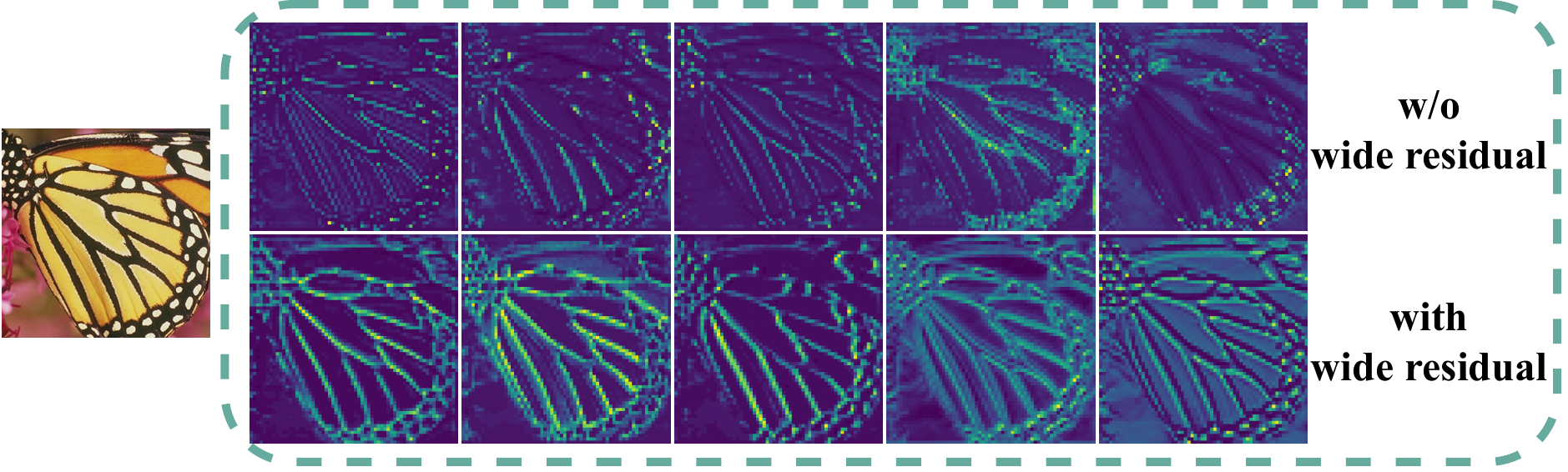}
\caption{Visualization of the effect of wide residual mechanism on extracted features.}
\label{RB_WRB_vision}
\end{figure}

\section{EXPERIMENTS}

\subsection{Datasets}
We use 800 pairs of HR-LR images from the DIV2K~\cite{timofte2017ntire} dataset for training, which includes images of various natural scenes. LR samples are generated using a bicubic downsampling method as used in~\cite{zhang2018image}. In addition, to evaluate the effectiveness of our method, we conducted tests on commonly used benchmark datasets, including Set5~\cite{bevilacqua2012low}, Set14~\cite{zeyde2010single}, BSDS100~\cite{martin2001database}, Urban100~\cite{huang2015single}, and Manga109~\cite{matsui2017sketch}.

\subsection{Implementation Details}
During training, we initialize the learning rate to 5e-4 and use the cosine annealing strategy to gradually decay it to 6.25e-6 over 1000 epochs. The optimizer used Adam, with the ${\beta _1}$ parameter set to 0.9 and the ${\beta _2}$ parameter set to 0.999. We randomly crop patches of size 48$ \times $48 from the training set as the input for training. Additionally, we apply data augmentation techniques such as random rotation and random flipping to these patches to enhance the dataset's variability. All our training is done using the Pytorch framework on an NVIDIA RTX 2080Ti. In the final model, the initial channel is set to 32 for the CNN part and 144 for the Transformer part. Furthermore, we apply weight normalization after the convolutional layers in the wide residual block to accelerate the convergence of training.

\subsection{Ablation Study}

\textbf{The effectiveness of WIRW and WCRW.} To assess the superiority of the residual blocks with wide activation mechanisms over normal residual blocks, we conducted experiments by replacing the WIRW and WCRW blocks with basic residual blocks as the baseline within the WDIB. We also investigated the impact of the number of channels before the activation function on the quantitative performance of SISR by setting the number of channels to 64 and 120, respectively. The results in Table~\ref{tab:RB} demonstrate the following observations: i) Our proposed FIWHN model achieves better performance and faster inference speed while using fewer parameters and Multi-adds compared to the baseline model. ii) Increasing the number of channels before the activation function leads to further improvements in the model performance with a slight increase in computational load. Furthermore, we provide visual results to illustrate the beneficial effects of the wide residual mechanism on feature extraction. In Figure~\ref{RB_WRB_vision}, we display the feature maps obtained by both normal residual units and our wide residual units. It is evident that the features extracted by the normal residual units lack many details in contour texture regions, which are crucial for accurate image recovery. In contrast, our proposed WIRW and WCRW units effectively mitigate the loss of these essential intermediate features mentioned above.


\begin{figure}[t]
\centering 
\includegraphics[width=0.42\textwidth,trim=0 0 20 0]{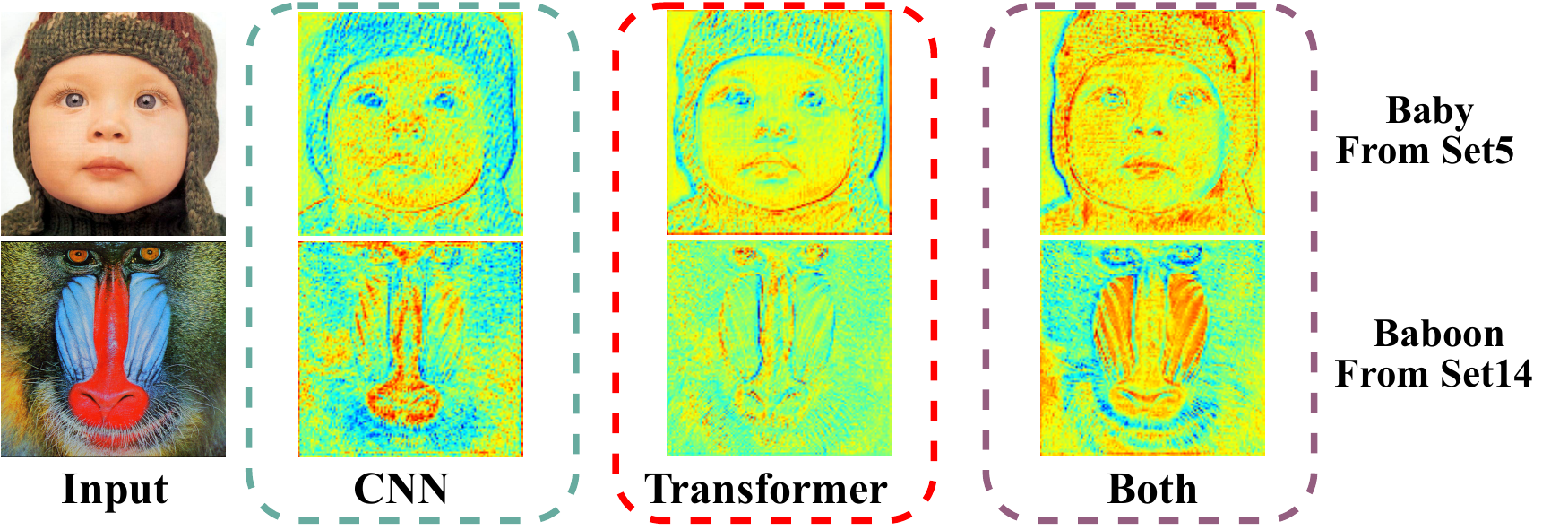}
\caption{Heat maps about the internal composition of FIWHN.}
\label{hotmap}
\end{figure}

\textbf{The effectiveness of WDIB.} Firstly, we analyze the internal components of WDIB in Table~\ref{tab:WRDC-SCF}, including our proposed WRDC (Case 1), SCF (Case 2), and $\otimes$ (adaptive multiplier, Case 4). By comparing case 1, case 2, and the baseline, we observe that our proposed WRDC module achieves slightly better performance than the baseline while saving approximately 13$\%$ of the number of parameters and 37$\%$ of the Multi-adds. The SCF module further improves the PSNR value by 0.18 dB with a parametric gain of less than 10K. The adaptive multiplier improves the PSNR by 0.04 dB compared to the baseline and does not introduce additional computational load or slow down the inference process. Overall, the integration of these submodules significantly enhances the performance.

Next, we provide a comparative analysis of our proposed WDIB with the sub-block of some methods in terms of PSNR values, model complexity, and inference speed in Table~\ref{tab:WDIB-Compare}. The compared methods include state-of-the-art models such as IMDN~\cite{hui2019lightweight}, RFDN~\cite{liu2020residual}, LatticeNet~\cite{luo2022lattice}, ESRT~\cite{lu2021efficient}, and LBNet~\cite{gao2022lightweight}. To ensure a fair comparison, we stack these blocks to achieve a similar number of parameters and then evaluate them comprehensively. As shown in the table, our proposed WDIB achieves the best performance with lower computational complexity. And our module has a shallower depth and faster inference speed compared to the other modules. Considering the trade-off between model size, inference speed, and reconstruction accuracy, WDIB proves to be a superior choice for efficient SISR.

\begin{table*}[t]
    \centering
    \small
    \caption{The PSNR/SSIM comparison with the state-of-the-art CNN-based SISR models. The best and the second-best results are highlighted and underlined, respectively.}
    \scalebox{0.99}{
    \begin{adjustbox}{width=1\linewidth}
    \begin{tabular}{|l|c|c|c|c|c|c|c|c|}
    \hline
    \multirow{2}{*}{Methods} & \multicolumn{1}{l|}{\multirow{2}{*}{Scale}} & \multirow{2}{*}{Params}& \multirow{2}{*}{Multi-adds} & Set5~\cite{bevilacqua2012low} & Set14~\cite{zeyde2010single} & BSDS100~\cite{martin2001database} & Urban100~\cite{huang2015single} & Manga109~\cite{matsui2017sketch} \\
    \cline{5-9} 
    & \multicolumn{1}{l|}{} & &  & PSNR/SSIM & PSNR/SSIM & PSNR/SSIM & PSNR/SSIM & PSNR/SSIM \\
    \hline
    \hline
    CARN~\cite{ahn2018fast}   & \multirow{14}{*}{$\times 2$}                           & 1592K  &222.8G   & 37.76/0.9590       & 33.52/0.9166           & 32.09/0.8978    & 31.92/0.9256     & 38.36/0.9765 \\
    IMDN~\cite{hui2019lightweight}                &       & 694K    &158.8G   & 38.00/0.9605       & 33.63/0.9177           &  32.19/0.8996    & 32.17/0.9283     & 38.88/0.9774 \\
    MADNet~\cite{lan2020madnet}                 &        & 878K    &187.1G   & 37.85/0.9600        & 33.38/0.9161            & 32.04/0.8979    & 31.62/0.9233     & - \\
    LAPAR-A~\cite{li2020lapar}                     &       & 548K    &171.0G   & 38.01/0.9605        & 33.62/0.9183            & 32.19/0.8999    & 32.10/0.9283   &38.67/0.9772  \\
    SMSR~\cite{wang2021exploring}                &       & 985K    &351.5G   & 38.00/0.9601       &33.64/0.9179            & 32.17/0.8990    & 32.19/0.9284     &38.76/0.9771 \\
    PFFN~\cite{zhang2021pffn}         		& 	 & 569K	& 138.3G	  & 38.07/0.9607       & 33.69/0.9192             & 32.21/0.8997   & 32.33/0.9298   &38.89/0.9775\\
    FDIWN~\cite{gao2022feature}         		& 	 & 629K	& 112.0G	  & 38.07/0.9608       & \underline{33.75}/\textbf{0.9201}            & 32.23/0.9003   & 32.40/0.9305   & 38.85/0.9774 \\
    LatticeNet-CL~\cite{luo2022lattice}           &       & 756K    &169.5G   & 38.09/0.9608        & 33.70/0.9188            & 32.21/0.9000    & 32.29/0.9291     & - \\
    FMEN~\cite{du2022fast}         		& 	 & 748K	& 172.0G	  & 38.10/0.9609       & \underline{33.75}/0.9192           & \underline{32.26}/0.9003   & \underline{32.41}/\underline{0.9311}   & \underline{38.95}/0.9778 \\
    HPUN-L~\cite{sun2023hybrid}         		& 	 & 714K	& 151.1G	  & 38.09/0.9608       & \textbf{33.79}/\underline{0.9198}           & 32.25/\underline{0.9006}   & 32.37/0.9307   & \textbf{39.07}/\underline{0.9779} \\
    GASSL-B~\cite{wang2023global}         		& 	 & 689K	& 158.2G	  & 38.08/0.9607       & \underline{33.75}/0.9194           & 32.24/0.9005   & 32.29/0.9298   & 38.92/0.9777 \\
    
    \hdashline[3pt/3pt]
    \textbf{FIWHN (Ours)}                               &      &705K     &137.7G  & \textbf{38.16}/\textbf{0.9613}   & 33.73/0.9194        & \textbf{32.27}/\textbf{0.9007}    &\textbf{32.75}/\textbf{0.9337}   &\textbf{39.07}/\textbf{0.9782}     \\

    \hline
    \hline
     
    CARN~\cite{ahn2018fast}   & \multirow{14}{*}{$\times 3$}                               & 1592K   &118.8G & 34.29/0.9255        &30.29/0.8407            & 29.06/0.8034    & 28.06/0.8493   & 33.43/0.9427    \\
    IMDN~\cite{hui2019lightweight}                 &        & 703K    &71.5G  & 34.36/0.9270       & 30.32/0.8417           & 29.09/0.8046    & 28.17/0.8519      & 33.61/0.9445 \\
    MADNet~\cite{lan2020madnet}                  &       & 930K     &88.4G  & 34.16/0.9253        & 30.21/0.8398           & 28.98/0.8023    & 27.77/0.8439           & -    \\
    LAPAR-A~\cite{li2020lapar}                      &       & 594K    &114.0G & 34.36/0.9267        & 30.34/0.8421            &29.11/0.8054    & 28.15/0.8523   & 33.51/0.9441   \\
    SMSR~\cite{wang2021exploring}                &       & 993K    &156.8G& 34.40/0.9270       &30.33/0.8412           & 29.10/0.8050    & 28.25/0.8536     & 33.68/0.9445\\
    PFFN~\cite{zhang2021pffn}                       &       & 558K    &69.1G & \underline{34.54}/\underline{0.9282}        & 30.42/0.8435          &29.17/0.8062    & 28.37/0.8566   & 33.63/0.9455   \\
    FDIWN~\cite{gao2022feature}                   &        & 645K    &51.5G  &34.52/0.9281         &30.42/0.8438          &29.14/0.8065   &28.36/0.8567     & 33.77/0.9456 \\
    LatticeNet-CL~\cite{luo2022lattice}           &       & 765K    &76.3G   & 34.46/0.9275        & 30.37/0.8422            & 29.12/0.8054    & 28.23/0.8525     & - \\
    FMEN~\cite{du2022fast}                   &        & 757K    &77.2G  &34.45/0.9275         &30.40/0.8435          &29.17/0.8063   &28.33/0.8562     & 33.86/0.9462 \\
    HPUN-L~\cite{sun2023hybrid}                   &        & 723K    &69.3G  &\textbf{34.56}/0.9281         &\underline{30.45}/\underline{0.8445}         &\underline{29.18}/\underline{0.8072}   &\underline{28.37}/\underline{0.8572}     & \underline{33.90}/\underline{0.9463} \\
    DDistill-SR~\cite{wang2023ddistill}         		& 	 & 665K	& 60.1G	  &34.43/0.9276         &30.39/0.8432          &29.16/0.8070   &28.31/0.8546     & \textbf{33.97}/0.9465 \\
    GASSL-B~\cite{wang2023global}         		& 	 & 691K	& 70.4G	  &34.47/0.9278         &30.39/0.8430          &29.15/0.8063   &28.27/0.8546     & 33.77/0.9455 \\
    
    \hdashline[3pt/3pt]
    \textbf{FIWHN (Ours)}                              &        &713K     &62.0G  &  34.50/ \textbf{0.9283}  & \textbf{30.50}/ \textbf{0.8451}   & \textbf{29.19}/ \textbf{0.8077}    & \textbf{28.62}/ \textbf{0.8607}    & \textbf{33.97}/\textbf{0.9472}  \\

    \hline
    \hline
    CARN~\cite{ahn2018fast}   & \multirow{14}{*}{$\times 4$}                             & 1592K  & 90.9G & 32.13/0.8937        & 28.60/0.7806            & 27.58/0.7349    & 26.07/0.7837   & 30.42/0.9070    \\
    IMDN~\cite{hui2019lightweight}                 &        & 715K   &40.9G  & 32.21/0.8948       & 28.58/0.7811          & 27.56/0.7353      & 26.04/0.7838      &30.45/0.9075 \\
    MADNet~\cite{lan2020madnet}                 &         & 1002K &54.1G & 31.95/0.8917       & 28.44/0.7780            & 27.47/0.7327    & 25.76/0.7746   & -    \\
    LAPAR-A~\cite{li2020lapar}                    &        & 659K   &94.0G  &32.15/0.8944        &28.61/0.7818            &27.61/0.7366    & 26.14/0.7871   &30.42/0.9074    \\
    SMSR~\cite{wang2021exploring}                &       & 1006K&89.1G& 32.12/0.8932          &28.55/0.7808           &27.55/0.7351    &26.11/0.7868     & 30.54/0.9085\\
    PFFN~\cite{zhang2021pffn}                     &        & 569K  &45.1G  &\textbf{32.36}/\textbf{0.8967}        &28.68/0.7827            &27.63/0.7370    & 26.26/0.7904   &30.50/0.9100    \\
    FDIWN~\cite{gao2022feature}                &       & 664K  &28.4G   & 32.23/0.8955   &28.66/0.7829            &27.62/0.7380        &26.28/0.7919                 & 30.63/0.9098 \\
    LatticeNet-CL~\cite{luo2022lattice}          &       & 777K  &43.6G   & 32.30/0.8958   &28.65/0.7822            &27.59/0.7365        &26.19/0.7855                 & - \\
    FMEN~\cite{du2022fast}                &       & 769K  & 44.2G   & 32.24/0.8955   &28.70/0.7839            &27.63/0.7379        &\underline{26.28}/0.7908                 & 30.70/0.9107 \\
    HPUN-L~\cite{sun2023hybrid}                &       & 734K  & 39.7G   & \underline{32.31}/\underline{0.8962}   &28.73/0.7842            &\underline{27.66}/0.7386        &26.27/\underline{0.7918}                & 30.77/0.9109 \\
    DDistill-SR~\cite{wang2023ddistill}         		& 	 & 675K	& 32.6G  & 32.29/0.8961   &28.69/0.7833            &27.65/0.7385        &26.25/0.7893                 & 30.79/0.9098 \\
    GASSL-B~\cite{wang2023global}         		& 	 & 694K  & 39.9G   & 32.27/0.8962   &\underline{28.74}/\textbf{0.7850}            &\underline{27.66}/\underline{0.7388}        &26.27/0.7914                 & \underline{30.92}/\underline{0.9122} \\
    
    \hdashline[3pt/3pt]
    \textbf{FIWHN (Ours)}                            &        &725K  &35.6G  & 32.30/\textbf{0.8967}    & \textbf{28.76}/\underline{0.7849}    & \textbf{27.68}/\textbf{0.7400}   &  \textbf{26.57}/\textbf{0.7989}       & \textbf{30.93}/\textbf{0.9131}     \\
    \hline
    \end{tabular}
    \end{adjustbox}
    }
    \label{tab:sota}
\end{table*}

\begin{figure}[t]
\centering
\begin{overpic}[width=0.96\linewidth]{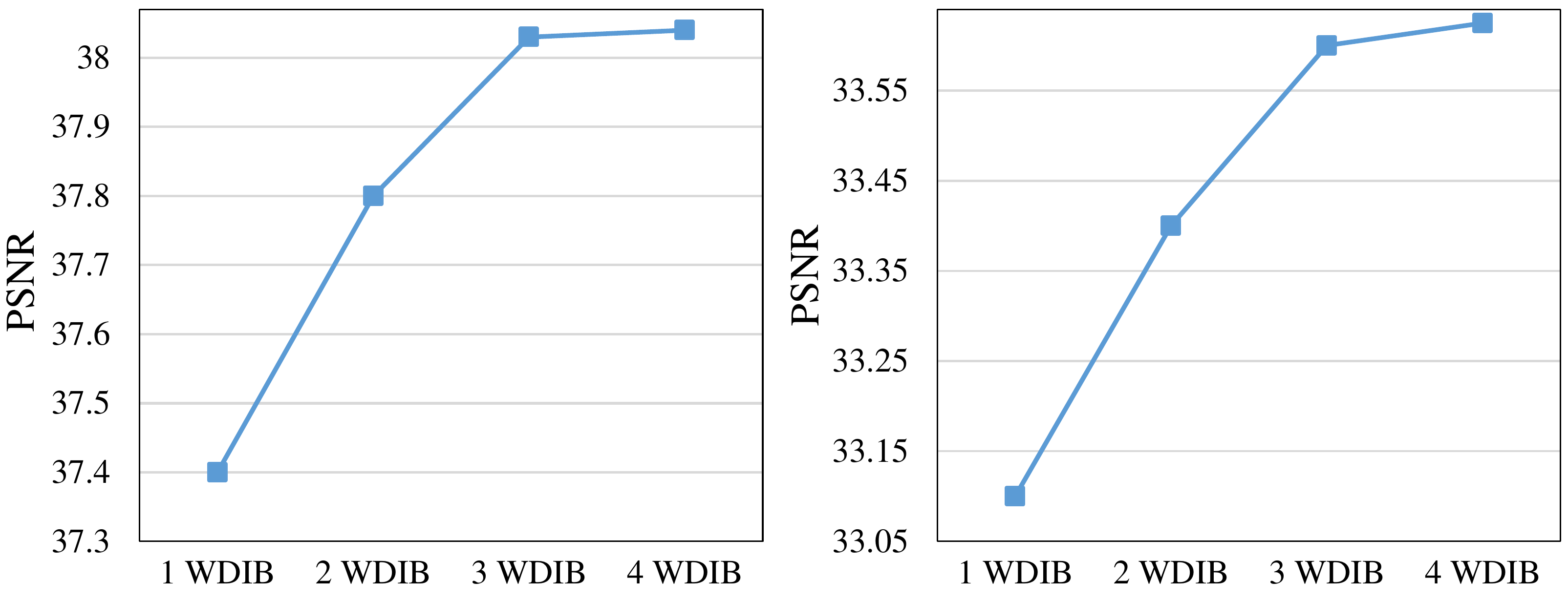}
\put(4.1,-5){\color{black}{\fontsize{8pt}{1pt}\selectfont (a) Ablation model on Set5 ($\times 2$)}}
\put(54,-5){\color{black}{\fontsize{8pt}{1pt}\selectfont (b) Ablation model on Set14 ($\times 2$)}}
\end{overpic}
\vspace{0.5cm}
\caption{Study on different numbers of WDIBs.}
\label{three-PSNR}
\vspace{-0.5cm}
\end{figure}

\textbf{The combination structure of FIWHN.} 
The combination structure of our proposed model consists of two main parts. The first is the feature grouping shuffle fusion part of the combined WDIB, which specifically addresses the issue of inadequate communication between blocks. In Table~\ref{tab:WRDC-SCF}, we compare the performance of the model with and without Block Interaction (BI) between blocks. Case 3, which includes the BI part, only has one additional BI compared to the baseline and almost no increase in computational load. The PSNR value of the model improves by 0.05 dB, indicating that communication between blocks benefits image reconstruction. Moreover, we also investigate the optimal number of inter-block feature mixing and fusion. As shown in Figure~\ref{three-PSNR}, the model achieves its best efficiency when the number of WDIBs forming FSWG is 3. Therefore, we use three WDIBs to form the FSWG.

\begin{table*}
	\centering
	\caption{Comparison with existing Transformer-base methods for $\times 4$ SR. }
	\label{tab:swinir}
	\scalebox{0.98}{
	\begin{tabular}{|c|c|c|c|c|c|c|c|c|c|}
		\hline
		\multirow{2}{*}{Methods} & \multirow{2}{*}{Params} & \multirow{2}{*}{Multi-adds} & \multirow{2}{*}{GPU}  & \multirow{2}{*}{Time}  & Set14~\cite{zeyde2010single} & BSDS100~\cite{martin2001database} & Urban100~\cite{huang2015single} & Manga109~\cite{matsui2017sketch} & Average  \\
		\cline{6-10}
		& & & & & PSNR / SSIM & PSNR / SSIM & PSNR / SSIM & PSNR / SSIM & PSNR / SSIM \\
		\hline
		\hline
		SwinIR-light~\cite{liang2021swinir} & 897K & 49.6G &10.5G & 55ms  & \underline{28.77} / \underline{0.7858} &  \textbf{27.69 / 0.7406} & \underline{26.47} / \underline{0.7980} & \underline{30.92} / \textbf{0.9151} & \underline{28.46} / \textbf{0.8192}\\
		LBNet~\cite{gao2022lightweight} & 742K & 38.9G & 6.4G & 49ms   & 28.68 / 0.7832 &  27.62 / 0.7382 & 26.27 / 0.7906 & 30.76 / 0.9111 & 28.30 / 0.8147\\
        CFIN~\cite{li2023cross} & 699K & 31.2G & 11.5G & 45ms   & 28.74 / 0.7849 &  27.68 / 0.7396 & 26.39 / 0.7946 & 30.73 / 0.9124 & 28.35 / 0.8169\\
        NGSwin~\cite{choi2023n} & 1019K & 36.4G &$>$12G &67ms   & \textbf{28.78} / \textbf{0.7859} &  27.66 / 0.7396 & 26.45 / 0.7963 & 30.80 / 0.9128 & 28.39 / 0.8184\\
        \hdashline[3pt/3pt]
		\textbf{FIWHN (Ours)} & 725K & 35.6G &7.5G & \textbf{38ms}  & 28.76 / 0.7849 &  \underline{27.68 / 0.7400} & \textbf{26.57} / \textbf{0.7989} & \textbf{30.93} / \underline{0.9131} & \textbf{28.49} / \underline{0.8186}\\
		\hline
	\end{tabular}
	}
\end{table*}

\begin{figure*}[htpb]
	\scriptsize
	\centering
	\scalebox{0.88}{
		\begin{tabular}{lc}
            
            \begin{adjustbox}{valign=t}
				\begin{tabular}{c}
					\includegraphics[width=0.22\textwidth, height=0.145\textheight]{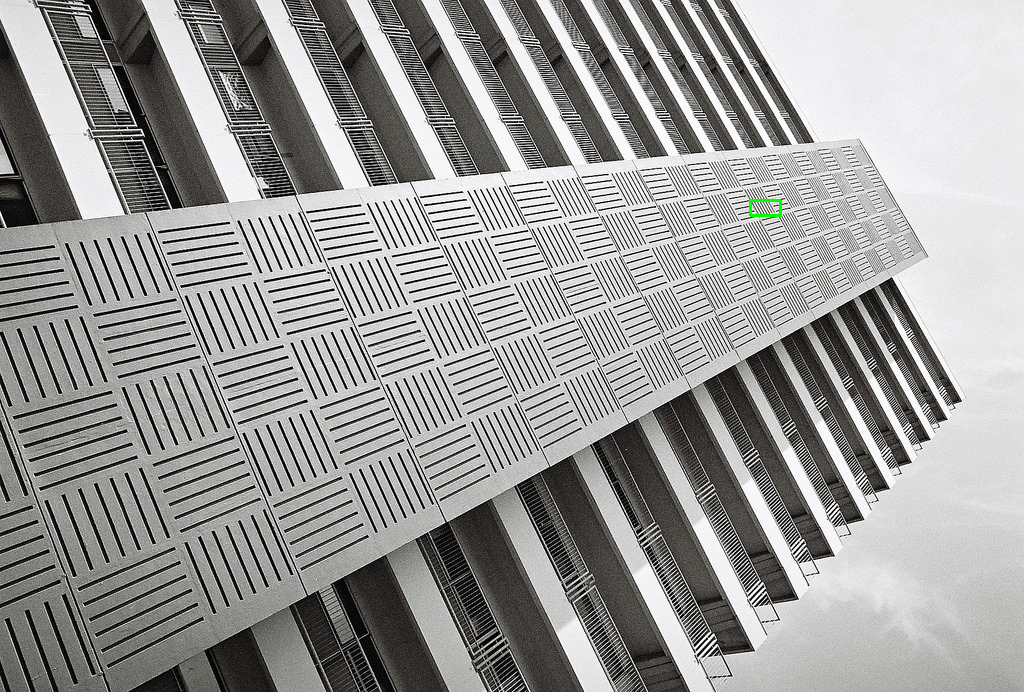} \\
					Urban100 ($\times 2$): \\
					img092 \\
				\end{tabular}
			\end{adjustbox}
			\hspace{-3mm}
			\begin{adjustbox}{valign=t}
				\begin{tabular}{cccccc}
					\includegraphics[width=0.135\textwidth, height=0.06\textheight]{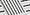} & 
					\hspace{-3mm}
					\includegraphics[width=0.135\textwidth, height=0.06\textheight]{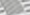} & 
					\hspace{-3mm}
					\includegraphics[width=0.135\textwidth, height=0.06\textheight]{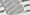} & 
                    \hspace{-3mm}
                    \includegraphics[width=0.135\textwidth, height=0.06\textheight]{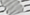} & 
                    \hspace{-3mm}
                    \includegraphics[width=0.135\textwidth, height=0.06\textheight]{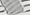} & 
					\hspace{-3mm}
					\includegraphics[width=0.135\textwidth, height=0.06\textheight]{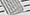} \\
					HR & \hspace{-2mm}
				  Bicubic & \hspace{-2mm}
					VDSR~\cite{kim2016accurate} & \hspace{-2mm}
                    CARN-M~\cite{ahn2018fast} & \hspace{-2mm}
                    CARN~\cite{ahn2018fast} & \hspace{-2mm}
					IMDN~\cite{hui2019lightweight} \\
					PSNR/SSIM & \hspace{-2mm}
					19.15/0.6690 & \hspace{-2mm}
					22.37/0.8185 & \hspace{-2mm}
                    22.54/0.8261 & \hspace{-2mm}
					23.21/0.8410& \hspace{-2mm}
					23.72/0.8511 \\
					\includegraphics[width=0.135\textwidth, height=0.06\textheight]{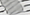} & 
					\hspace{-3mm}
					\includegraphics[width=0.135\textwidth, height=0.06\textheight]{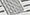} & 
					\hspace{-3mm}
                    \includegraphics[width=0.135\textwidth, height=0.06\textheight]{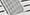} & 
                    \hspace{-3mm}
                    \includegraphics[width=0.135\textwidth, height=0.06\textheight]{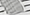} & 
                    \hspace{-3mm}
                    \includegraphics[width=0.135\textwidth, height=0.06\textheight]{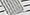} & 
					\hspace{-3mm}
					\includegraphics[width=0.135\textwidth, height=0.06\textheight]{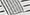} \\
					MADNet~\cite{lan2020madnet} & \hspace{-2mm}
				  AWSRN-M~\cite{wang2019lightweight} & \hspace{-2mm}
                    LBNet~\cite{gao2022lightweight} & \hspace{-2mm}
                    BSRN~\cite{li2022blueprint} & \hspace{-2mm}
                    FDIWN~\cite{gao2022feature} & \hspace{-2mm}
					\textbf{Ours} \\
					22.89/0.8337 & \hspace{-2mm}
					23.70/0.8510 & \hspace{-2mm}
                    24.05/0.8552 & \hspace{-2mm}
                    23.53/0.8498& \hspace{-2mm}
					23.84/0.8532& \hspace{-2mm}
					\textbf{24.40/0.8625} \\
				\end{tabular}
			\end{adjustbox}
             \\

             \begin{adjustbox}{valign=t}
				\begin{tabular}{c}
					\includegraphics[width=0.22\textwidth, height=0.145\textheight]{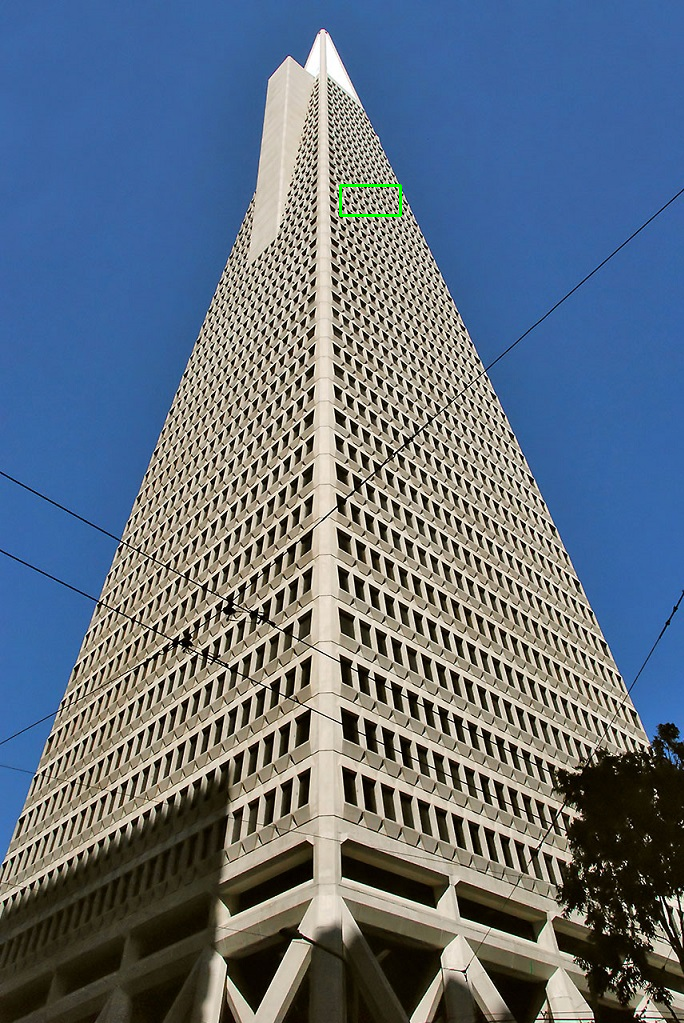} \\
					Urban100 ($\times 3$): \\
					img048 \\
				\end{tabular}
			\end{adjustbox}
			\hspace{-3mm}
			\begin{adjustbox}{valign=t}
				\begin{tabular}{cccccc}
					\includegraphics[width=0.135\textwidth, height=0.06\textheight]{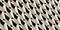} & 
					\hspace{-3mm}
					\includegraphics[width=0.135\textwidth, height=0.06\textheight]{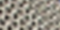} & 
					\hspace{-3mm}
					\includegraphics[width=0.135\textwidth, height=0.06\textheight]{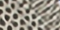} & 
                    \hspace{-3mm}
                    \includegraphics[width=0.135\textwidth, height=0.06\textheight]{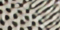} & 
                    \hspace{-3mm}
                    \includegraphics[width=0.135\textwidth, height=0.06\textheight]{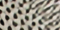} & 
					\hspace{-3mm}
					\includegraphics[width=0.135\textwidth, height=0.06\textheight]{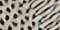} \\
					HR & \hspace{-2mm}
				  Bicubic & \hspace{-2mm}
					VDSR~\cite{kim2016accurate} & \hspace{-2mm}
                    CARN-M~\cite{ahn2018fast} & \hspace{-2mm}
                    CARN~\cite{ahn2018fast} & \hspace{-2mm}
					IMDN~\cite{hui2019lightweight} \\
					PSNR/SSIM & \hspace{-2mm}
					19.59/0.7581 & \hspace{-2mm}
					21.31/0.8550 & \hspace{-2mm}
                    21.97/0.8757 & \hspace{-2mm}
					22.32/0.8842 & \hspace{-2mm}
					22.36/0.8872 \\
					\includegraphics[width=0.135\textwidth, height=0.06\textheight]{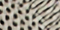} & 
					\hspace{-3mm}
					\includegraphics[width=0.135\textwidth, height=0.06\textheight]{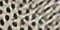} & 
					\hspace{-3mm}
					\includegraphics[width=0.135\textwidth, height=0.06\textheight]{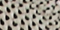} & 
                    \hspace{-3mm}
                    \includegraphics[width=0.135\textwidth, height=0.06\textheight]{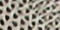} & 
                    \hspace{-3mm}
                    \includegraphics[width=0.135\textwidth, height=0.06\textheight]{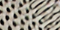} & 
					\hspace{-3mm}
					\includegraphics[width=0.135\textwidth, height=0.06\textheight]{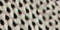} \\
					MADNet~\cite{lan2020madnet} & \hspace{-2mm}
				  AWSRN-M~\cite{wang2019lightweight} & 
                    \hspace{-2mm}
                    LBNet~\cite{gao2022lightweight} & \hspace{-2mm}
                    BSRN~\cite{li2022blueprint} & \hspace{-2mm}
                    FDIWN~\cite{gao2022feature} & \hspace{-2mm}
					\textbf{Ours} \\
					22.15/0.8784 & \hspace{-2mm}
					22.26/0.8860 & \hspace{-2mm}
                    22.84/0.8955 & \hspace{-2mm}
                    22.41/0.8871 & \hspace{-2mm}
					22.48/0.8892 & \hspace{-2mm}
					\textbf{23.49/0.9100} \\
				\end{tabular}
			\end{adjustbox}
             \\

             \begin{adjustbox}{valign=t}
				\begin{tabular}{c}
					\includegraphics[width=0.22\textwidth, height=0.145\textheight]{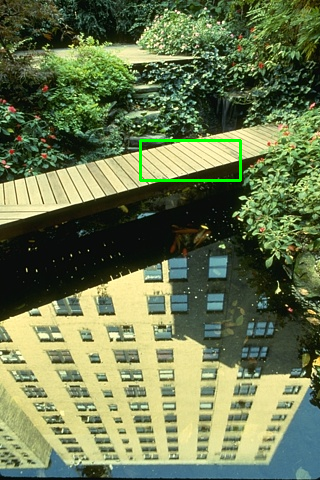} \\
					BSDS100 ($\times 4$): \\
					148026 \\
				\end{tabular}
			\end{adjustbox}
			\hspace{-3mm}
			\begin{adjustbox}{valign=t}
				\begin{tabular}{cccccc}
					\includegraphics[width=0.135\textwidth, height=0.06\textheight]{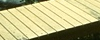} & 
					\hspace{-3mm}
					\includegraphics[width=0.135\textwidth, height=0.06\textheight]{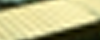} & 
					\hspace{-3mm}
					\includegraphics[width=0.135\textwidth, height=0.06\textheight]{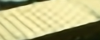} & 
                    \hspace{-3mm}
                    \includegraphics[width=0.135\textwidth, height=0.06\textheight]{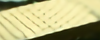} & 
                    \hspace{-3mm}
                    \includegraphics[width=0.135\textwidth, height=0.06\textheight]{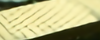} & 
					\hspace{-3mm}
					\includegraphics[width=0.135\textwidth, height=0.06\textheight]{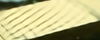} \\
					HR & \hspace{-2mm}
				  Bicubic & \hspace{-2mm}
					VDSR~\cite{kim2016accurate} & \hspace{-2mm}
                    CARN-M~\cite{ahn2018fast} & \hspace{-2mm}
                    CARN~\cite{ahn2018fast} & \hspace{-2mm}
					IMDN~\cite{hui2019lightweight} \\
					PSNR/SSIM & \hspace{-2mm}
					20.75/0.5573 & \hspace{-2mm}
					21.71/0.6578 & \hspace{-2mm}
                    21.94/0.6760 & \hspace{-2mm}
					22.10/0.6883 & \hspace{-2mm}
					22.13/0.6889 \\
					\includegraphics[width=0.135\textwidth, height=0.06\textheight]{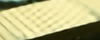} & 
					\hspace{-3mm}
					\includegraphics[width=0.135\textwidth, height=0.06\textheight]{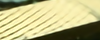} & 
					\hspace{-3mm}
					\includegraphics[width=0.135\textwidth, height=0.06\textheight]{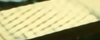} & 
                    \hspace{-3mm}
                    \includegraphics[width=0.135\textwidth, height=0.06\textheight]{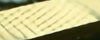} & 
                    \hspace{-3mm}
                    \includegraphics[width=0.135\textwidth, height=0.06\textheight]{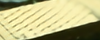} & 
					\hspace{-3mm}
					\includegraphics[width=0.135\textwidth, height=0.06\textheight]{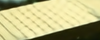} \\
					MADNet~\cite{lan2020madnet} & \hspace{-2mm}
				  AWSRN-M~\cite{wang2019lightweight} & 
                    \hspace{-2mm}
                    LBNet~\cite{gao2022lightweight} & \hspace{-2mm}
                    BSRN~\cite{li2022blueprint} & \hspace{-2mm}
                    FDIWN~\cite{gao2022feature} & \hspace{-2mm}
					\textbf{Ours} \\
					22.04/0.6823 & \hspace{-2mm}
					22.08/0.6880 & \hspace{-2mm}
                    22.24/0.6983 & \hspace{-2mm}
                    22.17/0.6930 & \hspace{-2mm}
					22.14/0.6940 & \hspace{-2mm}
					\textbf{22.33/0.7055} \\
				\end{tabular}
			\end{adjustbox}
             \\
            
			\begin{adjustbox}{valign=t}
				\begin{tabular}{c}
					\includegraphics[width=0.16\textwidth, height=0.12\textheight]{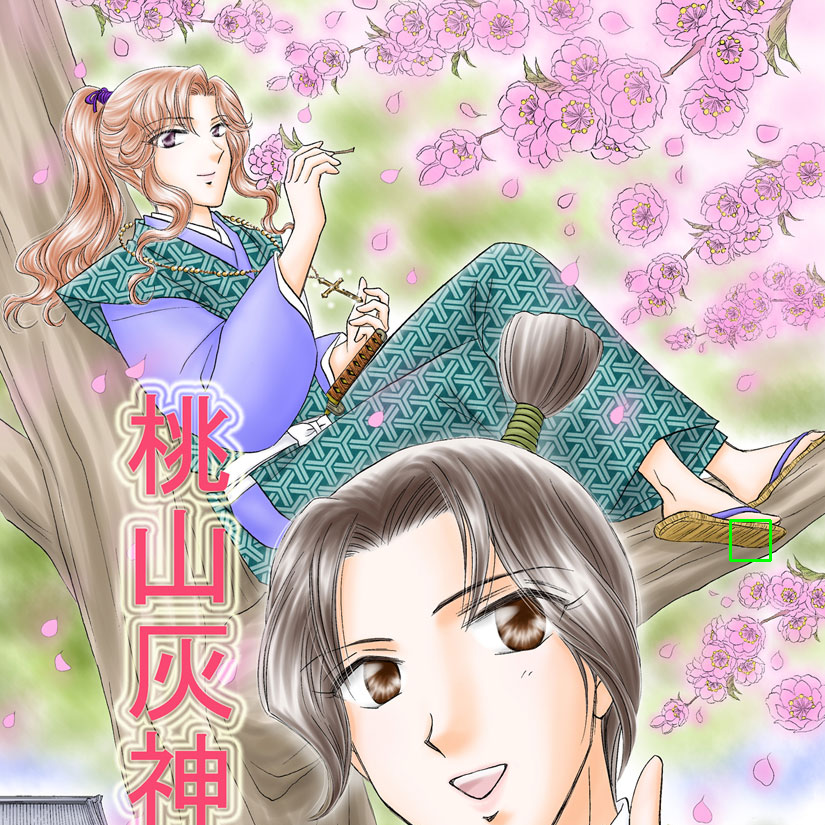} \\
					Manga109 ($\times 3$): \\
					MomoyamaHaikagura \\
				\end{tabular}
			\end{adjustbox}
			\hspace{-2mm}
			\begin{adjustbox}{valign=t}
				\begin{tabular}{cccc}
					\includegraphics[width=0.07\textwidth, height=0.05\textheight]{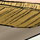} & 
					\hspace{-2mm}
					\includegraphics[width=0.07\textwidth, height=0.05\textheight]{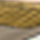} & 
					\hspace{-2mm}
					\includegraphics[width=0.07\textwidth, height=0.05\textheight]{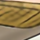} & 
					\hspace{-2mm}
					\includegraphics[width=0.07\textwidth, height=0.05\textheight]{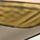} \\
					HR & \hspace{-2mm}
				  Bicubic & \hspace{-2mm}
					BSRN~\cite{li2022blueprint} & \hspace{-2mm}
					IMDN~\cite{hui2019lightweight} \\
					PSNR/SSIM & \hspace{-2mm}
					23.74/0.7034 & \hspace{-2mm}
					25.92/0.8550 & \hspace{-2mm}
					25.90/0.8534 \\
					\includegraphics[width=0.07\textwidth, height=0.05\textheight]{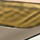} & 
					\hspace{-2mm}
					\includegraphics[width=0.07\textwidth, height=0.05\textheight]{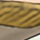} &
					\hspace{-2mm}
					\includegraphics[width=0.07\textwidth, height=0.05\textheight]{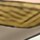} &
					\hspace{-2mm}
					\includegraphics[width=0.07\textwidth, height=0.05\textheight]{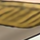} \\
					AWSRN-M~\cite{wang2019lightweight} & \hspace{-2mm}
					LBNet~\cite{gao2022feature} & \hspace{-2mm}
					FDIWN~\cite{gao2022lightweight} & \hspace{-2mm}
					\textbf{Ours} \\
					25.90/0.8534 & \hspace{-2mm}
					25.66/0.8435 & \hspace{-2mm}
					25.98/0.8557 & \hspace{-2mm}
					\textbf{26.16/0.8605} \\
				\end{tabular}
			\end{adjustbox}

            \begin{adjustbox}{valign=t}
				\begin{tabular}{c}
					\includegraphics[width=0.16\textwidth, height=0.12\textheight]{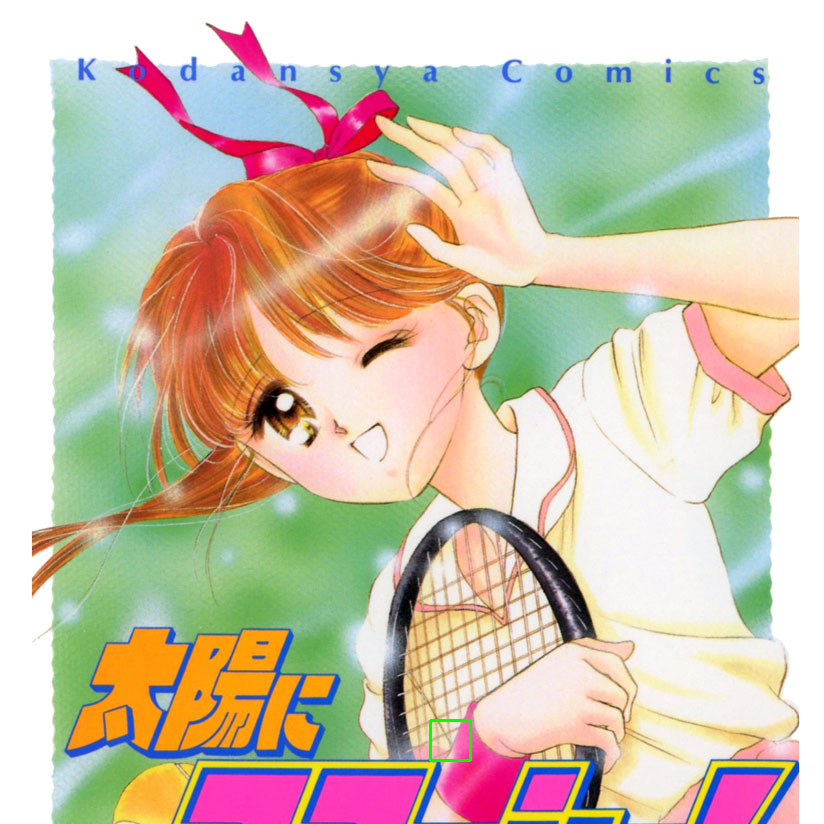} \\
					Manga109 ($\times 4$): \\
					TaiyouNiSmash \\
				\end{tabular}
			\end{adjustbox}
			\hspace{-3mm}
			\begin{adjustbox}{valign=t}
				\begin{tabular}{cccc}
					\includegraphics[width=0.07\textwidth, height=0.05\textheight]{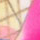} & 
					\hspace{-3mm}
					\includegraphics[width=0.07\textwidth, height=0.05\textheight]{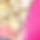} & 
					\hspace{-3mm}
					\includegraphics[width=0.07\textwidth, height=0.05\textheight]{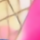} & 
					\hspace{-3mm}
					\includegraphics[width=0.07\textwidth, height=0.05\textheight]{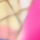} \\
					HR & \hspace{-3mm}
				  Bicubic & \hspace{-3mm}
					BSRN~\cite{li2022blueprint} & \hspace{-3mm}
					IMDN~\cite{hui2019lightweight} \\
					PSNR/SSIM & \hspace{-3mm}
					29.49/0.8712 & \hspace{-3mm}
					35.90/0.9536 & \hspace{-3mm}
					35.41/0.9499 \\
					\includegraphics[width=0.07\textwidth, height=0.05\textheight]{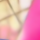} & 
					\hspace{-3mm}
					\includegraphics[width=0.07\textwidth, height=0.05\textheight]{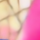} &
					\hspace{-3mm}
					\includegraphics[width=0.07\textwidth, height=0.05\textheight]{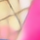} &
					\hspace{-3mm}
					\includegraphics[width=0.07\textwidth, height=0.05\textheight]{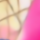} \\
					AWSRN-M~\cite{wang2019lightweight} & \hspace{-3mm}
					LBNet~\cite{gao2022feature} & \hspace{-3mm}
					FDIWN~\cite{gao2022lightweight} & \hspace{-3mm}
					\textbf{Ours} \\
					35.68/0.9513 & \hspace{-3mm}
					36.08/0.9522 & \hspace{-3mm}
					35.65/0.9508 & \hspace{-3mm}
					\textbf{36.50/0.9550} \\
				\end{tabular}
			\end{adjustbox}
			\\
			
	\end{tabular} }
	\vspace{3mm}
	\caption{Visual comparison of FIWHN with existing SISR methods.}
	\label{visual-comparison}
\end{figure*}

The second part of the combined structure involves the fusion of CNN and Transformer. In Table~\ref{tab:combine}, we evaluate different combinations as depicted in Figure~\ref{combine}. Our proposed scheme outperforms the implementations of simple CNN followed by Transformer, simple Transformer followed by CNN, and a simple parallel combination of CNN and Transformer. Importantly, this combination approach does not increase the overall computational load. These experiments on combination architectures demonstrate that an effective combination architecture can significantly enhance the model's representation ability. Furthermore, these results emphasize the importance of well-connecting and combining features from both local-based and global-based features in the middle layer to maximize the model's generalization ability. To visualize the effects of CNN and Transformer on the attention area, we provide the feature heat maps at different branches of the model in Figure~\ref{hotmap}. When the model contains only the CNN part, the attention is focused solely on the local area. Conversely, when the model contains only the Transformer part, it effectively captures global image information but may ignore certain local details. However, by integrating CNN and Transformer, the model can simultaneously consider both local and global areas, leading to the activation of more details.

\begin{table*}[t]
	\centering
	\small
	\caption{Comparison with existing SISR models on RealSR~\cite{cai2019toward} dataset.}
	\vspace{-0.2cm}
	\label{tab:realsr}
	\scalebox{0.9}{
	\begin{tabular}{|c|c|c|c|c|c|c|c|c|}
		\hline
		\multirow{2}{*}{Scale}  & Bicubic & SRCNN~\cite{dong2015image} & VDSR~\cite{dong2015image} & SRResNet~\cite{ledig2017photo} & IMDN~\cite{hui2019lightweight} & ESRT~\cite{lu2021efficient}& FDIWN~\cite{gao2022feature} &\textbf{FIWHN(ours)} \\ 
		\cline{2-9}
		& PSNR / SSIM & PSNR / SSIM & PSNR / SSIM & PSNR / SSIM & PSNR / SSIM & PSNR / SSIM  & PSNR / SSIM & PSNR / SSIM\\
		\hline
		\hline
		$ \times 2$    & 32.61 / 0.907 &  33.40 / 0.916 & 33.64 / 0.917 & 33.69 / 0.919 & 33.85 / 0.923 & 33.92 / 0.924 &33.68/0.9242  & \textbf{33.96 / 0.927}\\
		$ \times 3$    & 29.34 / 0.841 &  29.96 / 0.845 & 30.14 / 0.856 & 30.18 / 0.859 & 30.29 / 0.857 & 30.38 / 0.857 & 30.38 / 0.857 & \textbf{30.57 / 0.862}\\
		$ \times 4$     & 27.99 / 0.806 &  28.44 / 0.801 & 28.63 / 0.821 & 28.67 / 0.824 & 28.68 / 0.815 & 28.78 / 0.815 & 28.70 / 0.815 & \textbf{28.82 / 0.828}\\
		\hline
	\end{tabular}
	}
\end{table*}

\begin{figure*}[t]
	\scriptsize
	\centering
	\scalebox{0.87}{
		\begin{tabular}{lc}
            \begin{adjustbox}{valign=t}
				\begin{tabular}{c}
				\includegraphics[width=0.16\textwidth, height=0.12\textheight]{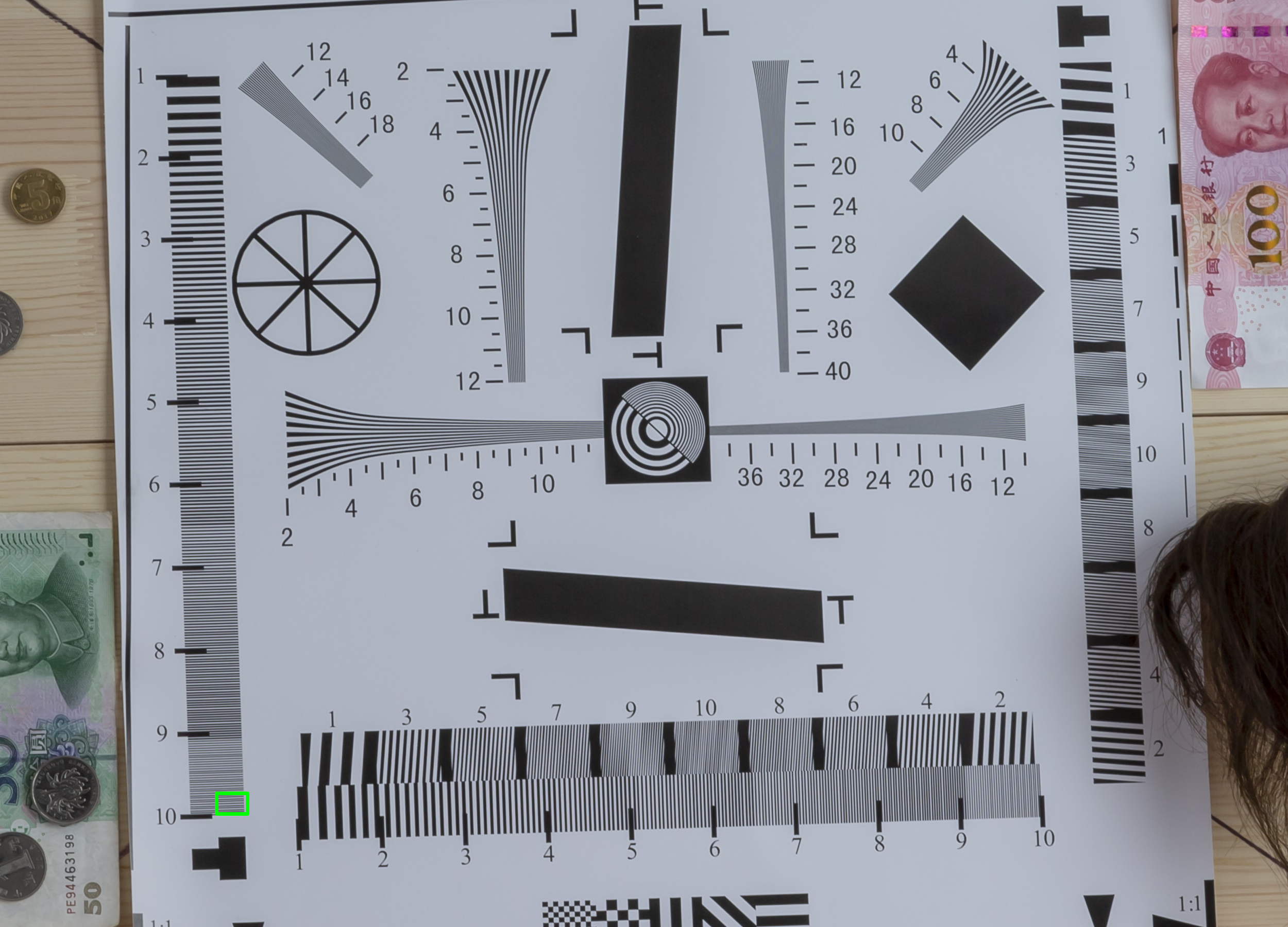} \\
					RealSR ($\times 2$): \\
					Canon010 \\
				\end{tabular}
			\end{adjustbox}
			\hspace{-3mm}
			\begin{adjustbox}{valign=t}
				\begin{tabular}{cc}
					\includegraphics[width=0.082\textwidth, height=0.045\textheight]{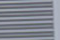} & 
					\hspace{-3mm}
					\includegraphics[width=0.082\textwidth, height=0.045\textheight]{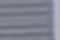} \\
					HR & \hspace{-3mm}
					LR \\
					PSNR/SSIM & \hspace{-3mm}
					29.23/0.9102 \\
					\includegraphics[width=0.082\textwidth, height=0.045\textheight]{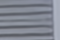} & 
					\hspace{-3mm}
					\includegraphics[width=0.082\textwidth, height=0.045\textheight]{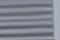} \\
					FDIWN~\cite{gao2022feature} & \hspace{-3mm}
					\textbf{Ours} \\
					32.24/0.9505 & \hspace{-3mm}
					\textbf{32.50/0.9517} \\
				\end{tabular}
			\end{adjustbox}
			
            \begin{adjustbox}{valign=t}
				\begin{tabular}{c}
					\includegraphics[width=0.16\textwidth, height=0.12\textheight]{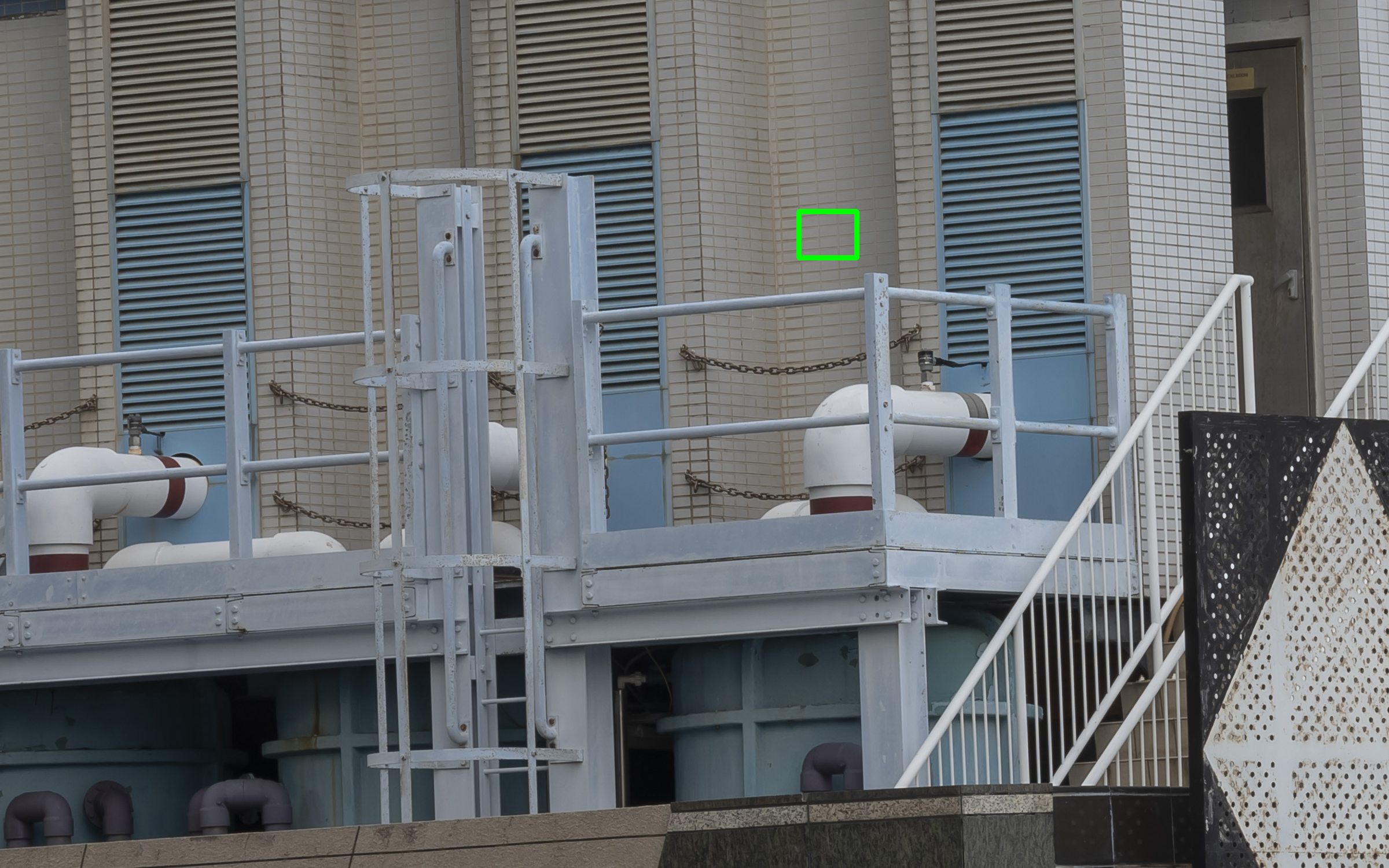} \\
					RealSR ($\times 3$): \\
					Nikon009 \\
				\end{tabular}
			\end{adjustbox}
			\hspace{-3mm}
			\begin{adjustbox}{valign=t}
				\begin{tabular}{cc}
					\includegraphics[width=0.082\textwidth, height=0.045\textheight]{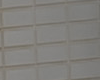} & 
					\hspace{-3mm}
					\includegraphics[width=0.082\textwidth, height=0.045\textheight]{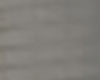} \\
					HR & \hspace{-3mm}
					LR \\
					PSNR/SSIM & \hspace{-3mm}
					29.33/0.8155 \\
					\includegraphics[width=0.082\textwidth, height=0.045\textheight]{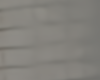} & 
					\hspace{-3mm}
					\includegraphics[width=0.082\textwidth, height=0.045\textheight]{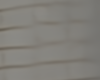} \\
					FDIWN~\cite{gao2022feature} & \hspace{-3mm}
					\textbf{Ours} \\
					31.05/0.8740 & \hspace{-3mm}
					\textbf{31.45/0.8802} \\
				\end{tabular}
			\end{adjustbox}

            \begin{adjustbox}{valign=t}
				\begin{tabular}{c}
					\includegraphics[width=0.16\textwidth, height=0.12\textheight]{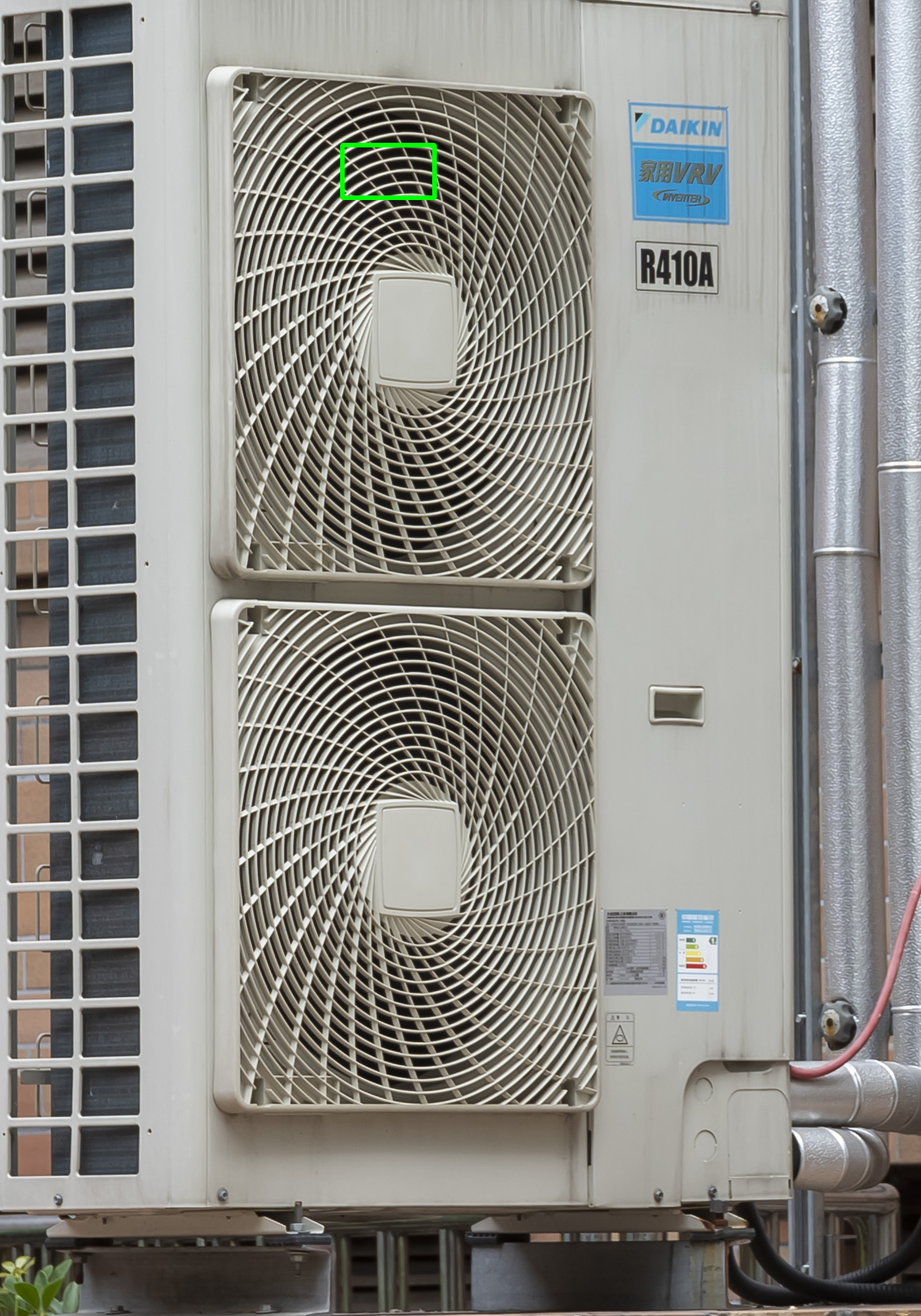} \\
					RealSR ($\times 4$): \\
					Nikon006 \\
				\end{tabular}
			\end{adjustbox}
			\hspace{-3mm}
			\begin{adjustbox}{valign=t}
				\begin{tabular}{cc}
					\includegraphics[width=0.082\textwidth, height=0.045\textheight]{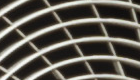} & 
					\hspace{-3mm}
					\includegraphics[width=0.082\textwidth, height=0.045\textheight]{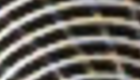} \\
					HR & \hspace{-3mm}
					LR \\
					PSNR/SSIM & \hspace{-3mm}
					24.03/0.7896 \\
					\includegraphics[width=0.082\textwidth, height=0.045\textheight]{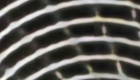} & 
					\hspace{-3mm}
					\includegraphics[width=0.082\textwidth, height=0.045\textheight]{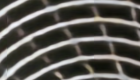} \\
					FDIWN~\cite{gao2022feature} & \hspace{-3mm}
					\textbf{Ours} \\
					24.77/0.8264 & \hspace{-3mm}
					\textbf{24.86/0.8327} \\
				\end{tabular}
			\end{adjustbox}
			\\
			
	\end{tabular} }
	\caption{Visual comparison on RealSR~\cite{cai2019toward} dataset (Including Nikon and Canon).}
	\label{realsr-img}
\end{figure*}

\textbf{Model complexity analysis.} 
In Figure~\ref{time_compare}, we present a comparison considering parameters, computational load, and inference time, in comparison to several state-of-the-art methods. The results demonstrate that our approach achieves superior performance across all scales while maintaining a reasonable number of parameters and computational effort. Notably, as shown in Figure~\ref{time_compare} (a), our method not only excels in performance but also outperforms most methods in terms of inference speed. Compared to the conference version FDIWN, FIWHN can achieve better performance using a shallower model depth due to the complementary global-local feature interaction, and therefore also significantly improves the inference speed. However, the inclusion of the highly computationally intensive Transformer also leads to an increase in model parameters and Muti-adds. It surpasses Transformer-based methods like LBNet \cite{gao2022lightweight}, CFIN \cite{li2023cross}, and NGramswin \cite{choi2023n}. This demonstrates that our method effectively strikes a superior balance among model complexity, performance, and inference speed when compared to alternative approaches. 

\subsection{Comparisons with State-of-the-Art Methods}
In this section, we provide a comprehensive comparison with state-of-the-art methods on benchmark datasets. The quantitative comparison results for $ \times 2$, $ \times 3$, and $ \times 4$ SISR are presented in Table~\ref{tab:sota}. It can be clearly seen that our FIWHN achieves the best performance across almost all datasets. Additionally, the number of parameters and the Multi-adds of our method is much lower than most of the methods. Furthermore, compared to our conference version FDIWN~\cite{gao2022feature}, we have further improved performance with only a marginal increase in computational cost. Notably, on Urban100 and Manga109 test sets, we observe an average performance gain of over 0.3 dB for all three scaling factors. These improvements clearly demonstrate the effectiveness of incorporating Transformers to complement the global features of the CNN model.

In addition, we also compare our method with several advanced Transformer-based methods in Table~\ref{tab:swinir}. Although our approach did not achieve optimal performance on some of the test sets, it can be clearly seen that our method outperforms LBNet~\cite{gao2022lightweight}, CFIN~\cite{li2023cross}, and NGSwin~\cite{choi2023n} on various datasets while being roughly comparable to SwinIR-light~\cite{liang2021swinir} in terms of average performance. However, it is worth noting that SwinIR-light utilizes additional pre-training strategies to enhance model performance and employs a larger patch size for training, which contributes to better performance. Moreover, SwinIR-light has significantly higher numbers of parameters and computations compared to our method. And training of our model can be efficiently conducted on an NVIDIA RTX 2080Ti, and our method demonstrates the fastest inference speed among these compared methods. Overall, the minute performance gap is challenging to discern visually, but our method's advantage of only about three-quarters of the computational consumption of SwinIR-light, and the inference time is faster, which is important for models running on devices with limited computational resources. These experiments collectively demonstrate that our FIWHN is a competitive method that excels in terms of performance, efficiency, and inference.

A qualitative comparison between our method and other methods is shown in Figure~\ref{visual-comparison}. To provide more convincing comparisons, we include the most recent CNN-based and Transformer-based methods. In addition to visual comparisons, we also provide the corresponding PSNR/SSIM values for a comprehensive analysis. As observed in this figure, our method consistently achieves higher PSNR and outperforms other methods in terms of visual quality, especially in capturing fine details across multiple validation sets.

\begin{figure*}[t]
\centering
\captionsetup{font=small}
\scriptsize
\hspace{0cm}
\scalebox{0.999}{
\begin{tabular}{c@{\extracolsep{0em}}@{\extracolsep{0.05em}}c@{\extracolsep{0.05em}}c@{\extracolsep{0.05em}}c@{\extracolsep{0.00em}}c@{\extracolsep{0.00em}}}
		\includegraphics[width=0.19\textwidth, height=0.08\textheight]{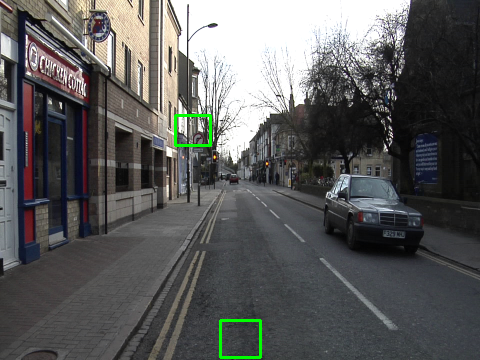}~~
		&\includegraphics[width=0.19\textwidth, height=0.08\textheight]{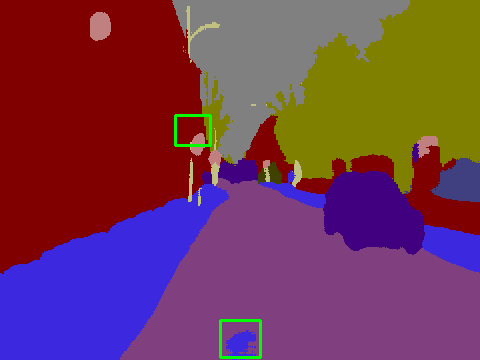}~~
		&\includegraphics[width=0.19\textwidth, height=0.08\textheight]{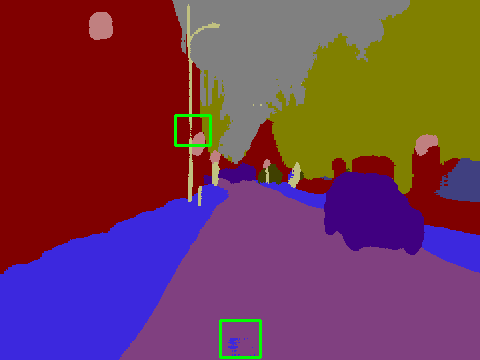}~~
        &\includegraphics[width=0.19\textwidth, height=0.08\textheight]{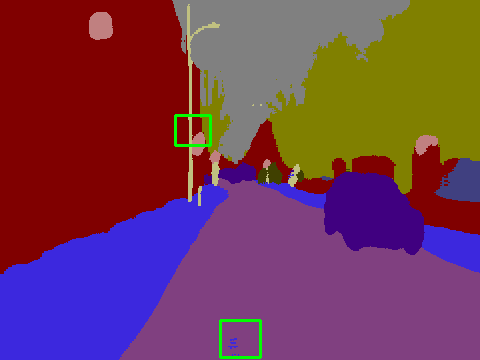}~~
        &\includegraphics[width=0.19\textwidth, height=0.08\textheight]{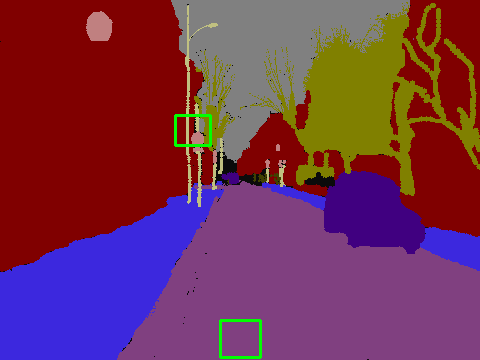}~~\\

        \includegraphics[width=0.188\textwidth, height=0.08\textheight]{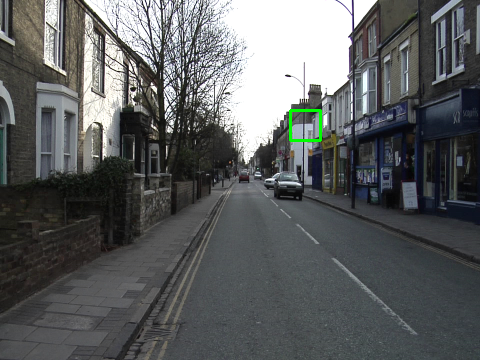}~~
		&\includegraphics[width=0.19\textwidth, height=0.08\textheight]{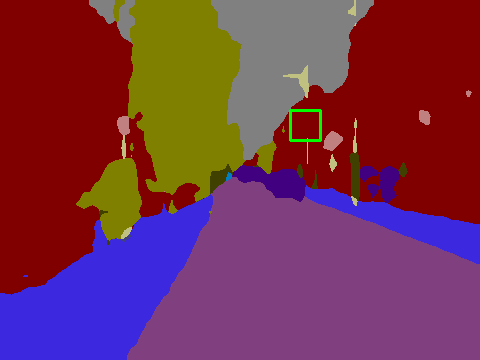}~~
		&\includegraphics[width=0.19\textwidth, height=0.08\textheight]{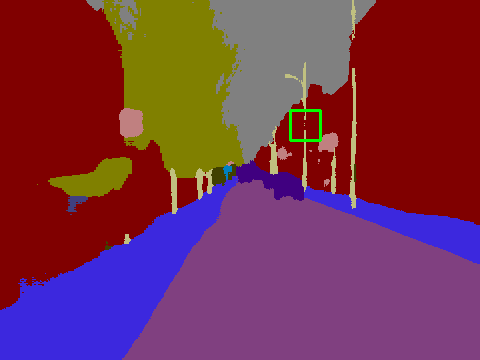}~~
        &\includegraphics[width=0.19\textwidth, height=0.08\textheight]{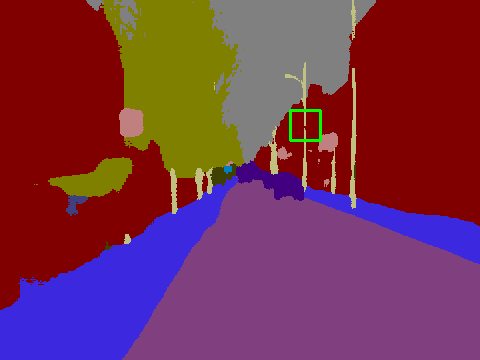}~~
        &\includegraphics[width=0.19\textwidth, height=0.08\textheight]{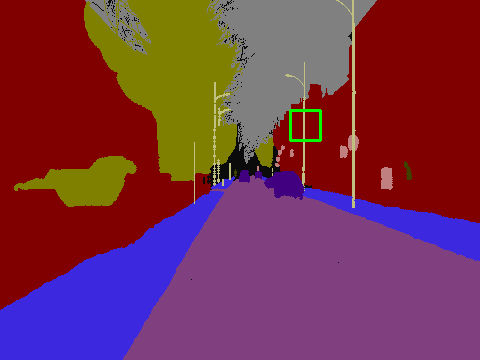}~~\\

 Street View & Bicubic & FDIWN~\cite{gao2022feature} & ~\textbf{Ours} & Ground Truth \\
\end{tabular}}
\caption{Comparison of qualitative image segmentation results ($\times 2$ SR) on CamVid~\cite{brostow2008segmentation} test set.}
\label{segmentation-result}
\end{figure*}

\subsection{Real-World Image Super-Resolution}
\textbf{Dataset and implement details.} 
For training and testing, we utilize the RealSR dataset~\cite{cai2019toward}. This dataset captures both HR and LR images in the same scene using the same camera but with different focal lengths. The degradation model in this dataset is more complex compared to the DIV2K dataset, and the degradation kernel varies spatially. It is worth noting that the dimensions of LR and HR images in this dataset are already aligned. Therefore, in our experiments, all methods remove the upsampling operation at the back end of the model. However, pixel alignment during image restoration becomes more challenging due to issues such as pixel drift and changes in scale factors resulting from adjusting the focal length. To alleviate the difficulty caused by pixel alignment, most methods adopt the strategy of dividing image patches into large patches when feeding them into the network during training. This operation helps mitigate the challenge of aligning pixels between numerous patches that do not effectively communicate with each other. Typically, These methods use image patches of size 128$\times $128 during training. However, for FDIWN and FIWHN, we set the patch to 64$\times $64 during training due to the higher training memory requirement associated with larger patch sizes. Therefore, our FIWHN can be trained on a single NVIDIA RTX 2080Ti GPU, while other methods cannot even be trained on two NVIDIA RTX 2080Ti GPUs due to their higher memory demands.

\textbf{Comparison results.}
The final quantitative comparison results are presented in Table~\ref{tab:realsr}. Despite facing a disadvantage in the training scenario, our method achieves the best results across all scales, particularly at a scale factor of $\times 3$, where our method outperforms the second-best method with a PSNR value of 0.19 dB. We also provide visual comparisons in Figure~\ref{realsr-img}, which highlight that our FIWHN can recover more textural details compared to the conference version FDIWN, resulting in restoration results that closely resemble the HR image. To further validate the effectiveness of FIWHN, we evaluate its benefits for street image semantic segmentation tasks. For this evaluation, we first downsample the images from the CamVid~\cite{brostow2008segmentation} dataset and then apply SISR methods to recover the high-quality images. Finally, we segment these images using the recently published real-time segmentation method FBSNet~\cite{gao2022fbsnet}. As shown in Figure~\ref{segmentation-result}, the segmentation results obtained from the images recovered by our method are closer to the ground truth. Notably, for segmented details such as the utility poles, our FIWHN significantly outperforms simple Bicubic and our conference version FDIWN. 



\section{CONCLUSIONS}
In this paper, we present a FIWHN that supports efficient SISR, which consists of groups of WDIB that are blended, fused, and weighted. The WDIB utilizes combinatorial coefficient learning to connect wide residual weighted units, which mitigates the loss of intermediate layers and induces different combinatorial structures. This enables features to exploit varying levels of information through residual connections and fusion. To further enhance the model, we explore a novel combined CNN and Transformer architecture to encourage capturing local and global feature interactions. Through extensive experiments, we demonstrate the efficacy of FIWHN in achieving efficient SISR and its applicability across diverse SISR scenarios and related downstream tasks.

\bibliographystyle{IEEEtran}
\bibliography{sample-base}

\begin{thebibliography}{10}
\providecommand{\url}[1]{#1}
\csname url@samestyle\endcsname
\providecommand{\newblock}{\relax}
\providecommand{\bibinfo}[2]{#2}
\providecommand{\BIBentrySTDinterwordspacing}{\spaceskip=0pt\relax}
\providecommand{\BIBentryALTinterwordstretchfactor}{4}
\providecommand{\BIBentryALTinterwordspacing}{\spaceskip=\fontdimen2\font plus
\BIBentryALTinterwordstretchfactor\fontdimen3\font minus \fontdimen4\font\relax}
\providecommand{\BIBforeignlanguage}[2]{{%
\expandafter\ifx\csname l@#1\endcsname\relax
\typeout{** WARNING: IEEEtran.bst: No hyphenation pattern has been}%
\typeout{** loaded for the language `#1'. Using the pattern for}%
\typeout{** the default language instead.}%
\else
\language=\csname l@#1\endcsname
\fi
#2}}
\providecommand{\BIBdecl}{\relax}
\BIBdecl

\bibitem{gao2022context}
G.~Gao, Y.~Yu, H.~Lu, J.~Yang, and D.~Yue, ``Context-patch representation learning with adaptive neighbor embedding for robust face image super-resolution,'' \emph{IEEE Transactions on Multimedia}, vol.~25, pp. 1879--1889, 2023.

\bibitem{dong2015image}
C.~Dong, C.~C. Loy, K.~He, and X.~Tang, ``Image super-resolution using deep convolutional networks,'' \emph{IEEE Transactions on Pattern Analysis and Machine Intelligence}, vol.~38, no.~2, pp. 295--307, 2015.

\bibitem{kim2016accurate}
J.~Kim, J.~K. Lee, and K.~M. Lee, ``Accurate image super-resolution using very deep convolutional networks,'' in \emph{CVPR}, 2016, pp. 1646--1654.

\bibitem{zhang2018image}
Y.~Zhang, K.~Li, K.~Li, L.~Wang, B.~Zhong, and Y.~Fu, ``Image super-resolution using very deep residual channel attention networks,'' in \emph{ECCV}, 2018, pp. 286--301.

\bibitem{gao2022feature}
G.~Gao, W.~Li, J.~Li, F.~Wu, H.~Lu, and Y.~Yu, ``Feature distillation interaction weighting network for lightweight image super-resolution,'' in \emph{AAAI}, vol.~36, no.~1, 2022, pp. 661--669.

\bibitem{zhang2021pffn}
D.~Zhang, C.~Li, N.~Xie, G.~Wang, and J.~Shao, ``Pffn: Progressive feature fusion network for lightweight image super-resolution,'' in \emph{ACMMM}, 2021, pp. 3682--3690.

\bibitem{chu2021fast}
X.~Chu, B.~Zhang, H.~Ma, R.~Xu, and Q.~Li, ``Fast, accurate and lightweight super-resolution with neural architecture search,'' in \emph{ICPR}.\hskip 1em plus 0.5em minus 0.4em\relax IEEE, 2021, pp. 59--64.

\bibitem{ahn2018fast}
N.~Ahn, B.~Kang, and K.-A. Sohn, ``Fast, accurate, and lightweight super-resolution with cascading residual network,'' in \emph{ECCV}, 2018, pp. 252--268.

\bibitem{sandler2018mobilenetv2}
M.~Sandler, A.~Howard, M.~Zhu, A.~Zhmoginov, and L.-C. Chen, ``Mobilenetv2: Inverted residuals and linear bottlenecks,'' in \emph{CVPR}, 2018, pp. 4510--4520.

\bibitem{luo2022lattice}
X.~Luo, Y.~Qu, Y.~Xie, Y.~Zhang, C.~Li, and Y.~Fu, ``Lattice network for lightweight image restoration,'' \emph{IEEE Transactions on Pattern Analysis and Machine Intelligence}, vol.~45, no.~4, pp. 4826--4842, 2023.

\bibitem{liang2021swinir}
J.~Liang, J.~Cao, G.~Sun, K.~Zhang, L.~Van~Gool, and R.~Timofte, ``Swinir: Image restoration using swin transformer,'' in \emph{ICCVW}, 2021, pp. 1833--1844.

\bibitem{lu2021efficient}
Z.~Lu, J.~Li, H.~Liu, C.~Huang, L.~Zhang, and T.~Zeng, ``Transformer for single image super-resolution,'' in \emph{CVPRW}, 2022, pp. 457--466.

\bibitem{gao2022lightweight}
G.~Gao, Z.~Wang, J.~Li, W.~Li, Y.~Yu, and T.~Zeng, ``Lightweight bimodal network for single-image super-resolution via symmetric cnn and recursive transformer,'' in \emph{IJCAI}, 2022, pp. 661--669.

\bibitem{zhang2022efficient}
X.~Zhang, H.~Zeng, S.~Guo, and L.~Zhang, ``Efficient long-range attention network for image super-resolution,'' in \emph{ECCV}, 2022, pp. 649--667.

\bibitem{li2023cross}
W.~Li, J.~Li, G.~Gao, W.~Deng, J.~Zhou, J.~Yang, and G.-J. Qi, ``Cross-receptive focused inference network for lightweight image super-resolution,'' \emph{IEEE Transactions on Multimedia}, vol.~26, pp. 864--877, 2024.

\bibitem{hui2019lightweight}
Z.~Hui, X.~Gao, Y.~Yang, and X.~Wang, ``Lightweight image super-resolution with information multi-distillation network,'' in \emph{ACMMM}, 2019, pp. 2024--2032.

\bibitem{li2020pams}
H.~Li, C.~Yan, S.~Lin, X.~Zheng, B.~Zhang, F.~Yang, and R.~Ji, ``Pams: Quantized super-resolution via parameterized max scale,'' in \emph{ECCV}.\hskip 1em plus 0.5em minus 0.4em\relax Springer, 2020, pp. 564--580.

\bibitem{lee2020learning}
W.~Lee, J.~Lee, D.~Kim, and B.~Ham, ``Learning with privileged information for efficient image super-resolution,'' in \emph{ECCV}.\hskip 1em plus 0.5em minus 0.4em\relax Springer, 2020, pp. 465--482.

\bibitem{wu2024transforming}
G.~Wu, J.~Jiang, J.~Jiang, and X.~Liu, ``Transforming image super-resolution: a convformer-based efficient approach,'' \emph{arXiv:2401.05633}, 2024.

\bibitem{yu2018wide}
J.~Yu, Y.~Fan, J.~Yang, N.~Xu, Z.~Wang, X.~Wang, and T.~Huang, ``Wide activation for efficient and accurate image super-resolution,'' \emph{arXiv preprint arXiv:1808.08718}, 2018.

\bibitem{bao2023sctanet}
Q.~Bao, Y.~Liu, B.~Gang, W.~Yang, and Q.~Liao, ``Sctanet: A spatial attention-guided cnn-transformer aggregation network for deep face image super-resolution,'' \emph{IEEE Transactions on Multimedia}, vol.~25, pp. 8554--8565, 2024.

\bibitem{qi2023efficient}
H.~Qi, Y.~Qiu, X.~Luo, and Z.~Jin, ``An efficient latent style guided transformer-cnn framework for face super-resolution,'' \emph{IEEE Transactions on Multimedia}, vol.~26, pp. 1589--1599, 2024.

\bibitem{choi2023n}
H.~Choi, J.~Lee, and J.~Yang, ``N-gram in swin transformers for efficient lightweight image super-resolution,'' in \emph{CVPR}, 2023, pp. 2071--2081.

\bibitem{zhang2018shufflenet}
X.~Zhang, X.~Zhou, M.~Lin, and J.~Sun, ``Shufflenet: An extremely efficient convolutional neural network for mobile devices,'' in \emph{CVPR}, 2018, pp. 6848--6856.

\bibitem{wang2022faceformer}
Y.~Wang, T.~Lu, Y.~Zhang, Z.~Wang, J.~Jiang, and Z.~Xiong, ``Faceformer: aggregating global and local representation for face hallucination,'' \emph{IEEE Transactions on Circuits and Systems for Video Technology}, 2022.

\bibitem{liu2020residual}
J.~Liu, J.~Tang, and G.~Wu, ``Residual feature distillation network for lightweight image super-resolution,'' in \emph{ECCV}, 2020, pp. 41--55.

\bibitem{timofte2017ntire}
R.~Timofte, E.~Agustsson, L.~Van~Gool, M.-H. Yang, and L.~Zhang, ``Ntire 2017 challenge on single image super-resolution: Methods and results,'' in \emph{CVPRW}, 2017, pp. 114--125.

\bibitem{bevilacqua2012low}
M.~Bevilacqua, A.~Roumy, C.~Guillemot, and M.~L. Alberi-Morel, ``Low-complexity single-image super-resolution based on nonnegative neighbor embedding,'' in \emph{BMVC}, 2012, pp. 135.1--135.10.

\bibitem{zeyde2010single}
R.~Zeyde, M.~Elad, and M.~Protter, ``On single image scale-up using sparse-representations,'' in \emph{ICCS}, 2010, pp. 711--730.

\bibitem{martin2001database}
D.~Martin, C.~Fowlkes, D.~Tal, and J.~Malik, ``A database of human segmented natural images and its application to evaluating segmentation algorithms and measuring ecological statistics,'' in \emph{ICCV}, 2001, pp. 416--423.

\bibitem{huang2015single}
J.-B. Huang, A.~Singh, and N.~Ahuja, ``Single image super-resolution from transformed self-exemplars,'' in \emph{CVPR}, 2015, pp. 5197--5206.

\bibitem{matsui2017sketch}
Y.~Matsui, K.~Ito, Y.~Aramaki, A.~Fujimoto, T.~Ogawa, T.~Yamasaki, and K.~Aizawa, ``Sketch-based manga retrieval using manga109 dataset,'' \emph{Multimedia Tools and Applications}, vol.~76, no.~20, pp. 21\,811--21\,838, 2017.

\bibitem{lan2020madnet}
R.~Lan, L.~Sun, Z.~Liu, H.~Lu, C.~Pang, and X.~Luo, ``Madnet: a fast and lightweight network for single-image super resolution,'' \emph{IEEE Transactions on Cybernetics}, vol.~51, no.~3, pp. 1443--1453, 2020.

\bibitem{li2020lapar}
W.~Li, K.~Zhou, L.~Qi, N.~Jiang, J.~Lu, and J.~Jia, ``Lapar: Linearly-assembled pixel-adaptive regression network for single image super-resolution and beyond,'' in \emph{NeurIPS}, vol.~33, 2020, pp. 20\,343--20\,355.

\bibitem{wang2021exploring}
L.~Wang, X.~Dong, Y.~Wang, X.~Ying, Z.~Lin, W.~An, and Y.~Guo, ``Exploring sparsity in image super-resolution for efficient inference,'' in \emph{CVPR}, 2021, pp. 4917--4926.

\bibitem{du2022fast}
Z.~Du, D.~Liu, J.~Liu, J.~Tang, G.~Wu, and L.~Fu, ``Fast and memory-efficient network towards efficient image super-resolution,'' in \emph{CVPRW}, 2022, pp. 853--862.

\bibitem{sun2023hybrid}
B.~Sun, Y.~Zhang, S.~Jiang, and Y.~Fu, ``Hybrid pixel-unshuffled network for lightweight image super-resolution,'' in \emph{AAAI}, vol.~37, no.~2, 2023, pp. 2375--2383.

\bibitem{wang2023global}
H.~Wang, Y.~Zhang, C.~Qin, L.~Van~Gool, and Y.~Fu, ``Global aligned structured sparsity learning for efficient image super-resolution,'' \emph{IEEE Transactions on Pattern Analysis and Machine Intelligence}, vol.~45, no.~9, pp. 10\,974--10\,989, 2023.

\bibitem{wang2023ddistill}
Y.~Wang, T.~Su, Y.~Li, J.~Cao, G.~Wang, and X.~Liu, ``Ddistill-sr: Reparameterized dynamic distillation network for lightweight image super-resolution,'' \emph{IEEE Transactions on Multimedia}, vol.~25, pp. 7222--7234, 2023.

\bibitem{wang2019lightweight}
C.~Wang, Z.~Li, and J.~Shi, ``Lightweight image super-resolution with adaptive weighted learning network,'' \emph{arXiv preprint arXiv:1904.02358}, 2019.

\bibitem{li2022blueprint}
Z.~Li, Y.~Liu, X.~Chen, H.~Cai, J.~Gu, Y.~Qiao, and C.~Dong, ``Blueprint separable residual network for efficient image super-resolution,'' in \emph{CVPR}, 2022, pp. 833--843.

\bibitem{cai2019toward}
J.~Cai, H.~Zeng, H.~Yong, Z.~Cao, and L.~Zhang, ``Toward real-world single image super-resolution: A new benchmark and a new model,'' in \emph{ICCV}, 2019, pp. 3086--3095.

\bibitem{ledig2017photo}
C.~Ledig, L.~Theis, F.~Husz{\'a}r, J.~Caballero, A.~Cunningham, A.~Acosta, A.~Aitken, A.~Tejani, J.~Totz, Z.~Wang \emph{et~al.}, ``Photo-realistic single image super-resolution using a generative adversarial network,'' in \emph{CVPR}, 2017, pp. 4681--4690.

\bibitem{brostow2008segmentation}
G.~J. Brostow, J.~Shotton, J.~Fauqueur, and R.~Cipolla, ``Segmentation and recognition using structure from motion point clouds,'' in \emph{ECCV}.\hskip 1em plus 0.5em minus 0.4em\relax Springer, 2008, pp. 44--57.

\bibitem{gao2022fbsnet}
G.~Gao, G.~Xu, J.~Li, Y.~Yu, H.~Lu, and J.~Yang, ``Fbsnet: A fast bilateral symmetrical network for real-time semantic segmentation,'' \emph{IEEE Transactions on Multimedia}, vol.~25, pp. 3273--3283, 2023.

\end{thebibliography}


\begin{IEEEbiographynophoto}{Wenjie Li}
received the M.S. degree with the College of Automation \& College of Artificial Intelligence, Nanjing University of Posts and Telecommunications. He is currently pursuing the Ph.D. degree with the School of Artificial Intelligence, Beijing University of Posts and Telecommunications. His research interests include image restoration.
\end{IEEEbiographynophoto}


\begin{IEEEbiographynophoto}{Juncheng Li}
received the Ph.D. degree in Computer Science and Technology from East China Normal University, in 2021. He is currently an Assistant Professor at the School of Communication \& Information Engineering, Shanghai University. His main research interests include image restoration, computer vision, and medical image processing.
\end{IEEEbiographynophoto}

\begin{IEEEbiographynophoto}{Guangwei Gao}
(Senior Member, IEEE) received the Ph.D. degree in pattern recognition and intelligence systems from Nanjing University of Science and Technology, Nanjing, in 2014. He is currently a Professor at Nanjing University of Posts and Telecommunications. His research interests include pattern recognition and computer vision. 
\end{IEEEbiographynophoto}

\begin{IEEEbiographynophoto}{Weihong Deng}
(Member, IEEE) received the Ph.D. degree in signal and information processing from the Beijing University of Posts and Telecommunications (BUPT), Beijing, China, in 2009. He was a professor at the School of Artificial Intelligence, BUPT. His research interests include trustworthy biometrics and affective computing, with a particular emphasis on face recognition and expression analysis.
\end{IEEEbiographynophoto}

\begin{IEEEbiographynophoto}{Jian Yang}
(Member, IEEE) received the Ph.D. degree from the Nanjing University of Science and Technology (NJUST), Nanjing, China. He is currently a professor with the School of Computer Science and Technology, NJUST. He is the author of more than 400 scientific papers in pattern recognition and computer vision. 
His research interests include pattern recognition, computer vision, and machine learning. He is/was an associate editor for Pattern Recognition and IEEE Transactions Neural Networks and Learning Systems. He is a fellow of IAPR.
\end{IEEEbiographynophoto}

\begin{IEEEbiographynophoto}{Guojun Qi}
(Fellow, IEEE) has been a faculty member with the Department of Computer Science, University of Central Florida, since August 2014. Since August 2018, he has been the Chief Scientist leading and overseeing the International Research and Development Team for multiple artificial intelligence services on the Huawei Cloud. He is currently a Professor and the Chief Scientist who oversees the Artificial Intelligence Research Center, Westlake University, and the OPPO U.S. Research Center. His research interests include machine learning and knowledge discovery from multi-modal data to build smart and reliable information and decision-making systems.
\end{IEEEbiographynophoto}

\begin{IEEEbiographynophoto}{Chia-Wen Lin}
(Fellow, IEEE) received the Ph.D. degree in electrical engineering from National Tsing Hua University (NTHU), Hsinchu, Taiwan, in 2000. He was with the Department of Computer Science and Information Engineering, National Chung Cheng University, Taiwan, from 2000 to 2007. He is currently a Professor with the Department of Electrical Engineering and the Institute of Communications Engineering, NTHU. He is also the Deputy Director of the AI Research Center, NTHU. His research interests include image and video processing, computer vision, and video networking. 
\end{IEEEbiographynophoto}


\end{document}